\documentclass[journal]{IEEEtran}

\usepackage{graphicx}
\usepackage{amsmath}
\usepackage{amssymb}
\usepackage{booktabs}
\usepackage{algorithm}
\usepackage{algorithmic}
\usepackage{csquotes}
\usepackage{rotating}
\usepackage{amsfonts}
\usepackage{array}
\usepackage{multirow}
\usepackage{tabularx}
\usepackage{lipsum} 
\usepackage{makecell}
\usepackage{adjustbox}
\usepackage{xspace}
\DeclareMathOperator*{\argmax}{arg\,max}
\DeclareMathOperator*{\argmin}{arg\,min}
\usepackage{hhline,array}
\usepackage{array, makecell, rotating}

\usepackage{newfloat}
\usepackage{listings}

\newcommand{\answerTODO}[1][]{\textcolor{red}{\bf [TODO]}}
\newcommand{\justificationTODO}[1][]{\textcolor{red}{\bf [TODO]}}
\usepackage{cleveref}
\crefname{section}{Sec.}{Secs.}
\Crefname{section}{Section}{Sections}
\crefname{table}{Table}{Tables}
\Crefname{table}{Table}{Tables}
\crefname{figure}{Figure}{Figures}
\crefname{figure}{Figure}{Figures}
\crefname{equation}{Equation}{Equations}
\crefname{equation}{Eq.}{Eqs.}
\crefname{algorithm}{Algorithm}{Algorithms}
\crefname{algorithm}{Algorithm}{Algorithms}

\usepackage{xcolor}

\usepackage{xspace}
\newcommand{\ie}{{\emph{i.e.}},\xspace}

\newcommand{\eg}{{\emph{e.g.}},\xspace}
\newcommand{\wrt}{{\emph{w.r.t.}}\xspace}

\newcommand{\etal}{{\emph{et al.}}}

\begin{document}

\title{A Generative Victim Model for Segmentation}
%
%
%

\author{Aixuan Li, Jing Zhang,~\IEEEmembership{Member,~IEEE,} Jiawei Shi,~\IEEEmembership{Member,~IEEE,}, Yiran Zhong, Yuchao Dai,~\IEEEmembership{Member,~IEEE}
}
\maketitle

\begin{abstract}
We find that the well-trained victim models~(VMs), against which the attacks are generated, serve as fundamental prerequisites for adversarial attacks, \ie a segmentation VM is needed to generate attacks for segmentation. In this context, the victim model is assumed to be robust to achieve effective adversarial perturbation generation.
Instead of focusing on improving the robustness of the task-specific victim models, we shift our attention to image generation.
From an image generation perspective, we derive a novel VM for segmentation, aiming to generate adversarial perturbations for segmentation tasks without requiring models explicitly designed for image segmentation. Our approach to adversarial attack generation diverges from conventional white-box or black-box attacks, offering a fresh outlook on adversarial attack strategies.
Experiments show that our attack method is able to generate effective adversarial attacks with good transferability.
\end{abstract}


%
\IEEEpeerreviewmaketitle

\section{Introduction}
\label{sec:intro}
\IEEEPARstart{E}{xisting} adversarial attacks can be broadly categorized into white-box attacks and black-box attacks. The former~\cite{gu2022segpgd,adversarial_training,arnab2018robustness,madry2017towards} possess comprehensive knowledge about the victim model (VM), \eg~model inputs, its architecture and weights. The latter~\cite{papernot2017practical,zhou2020dast,andriushchenko2020square,guo2019simple,al2019sign}, on the other hand, only have information about model inputs, and an oracle that enables querying for outputs.
Although white-box attacks are usually proven more effective,
their practical utility in real-world scenarios is restricted due to the challenge of accessing a model's internal parameters during deployment.
In contrast, query-based black-box attacks~\cite{brendel2018decision,ilyas2018black,uesato2018adversarial} generate adversarial perturbations by leveraging the victim model's predictions without necessitating access to model parameters, rendering them more convenient for real-world applications. However, these black-box attacks often require thousands of queries to generate effective perturbations, resulting in decreased efficiency. Moreover, the attacks may fail when the victim network changes~\cite{xie2017adversarial}, showing a lower degree of transferability.

 \begin{figure*}[!ht]
	\centering %
	\includegraphics[width=0.95\textwidth]{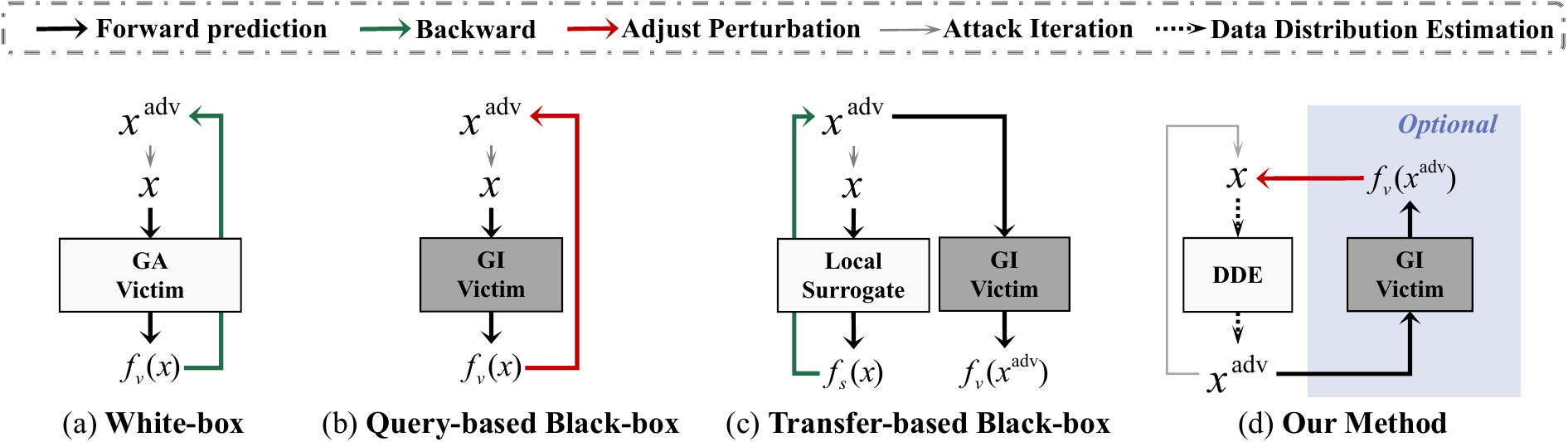} 
	\caption{\textit{Adversarial attacks comparison}. \textbf{GA} is Gradient-accessible, \textbf{GI} is Gradient-inaccessible, and \textbf{DDE} denotes Data Distribution Estimation Model. $f_v$ and $f_s$ denote the victim model and the local surrogate model, respectively. As shown in (d), our method can choose whether to query the victim model or not, which is quite flexible.} 
	\label{Fig.attack_diff} 
\end{figure*}

Considering that the victim models are usually not so readily accessible, transferable
black-box attacks~\cite{wu2020skip,wang2022generating,kim2022diverse,zhang2023transferable,ma2023transferable} are designed to achieve the trade-off, where adversarial examples are generated from a white-box surrogate model and then transferred to the target black-box model. However, the transferable black-box attack methods can not update the attacks based on the victim model, leading to less effective attacks. 
We show a visual comparison of each setting in  \cref{Fig.attack_diff}, which
clearly shows that
a task-specific victim model is assumed available for all the three settings, \eg~a classification model is needed for adversarial attacks for classification, and a segmentation model is assumed available to generate adversarial perturbation for segmentation.

For the first time, we reconsider the concept of victim models from the perspective of image generation, drawing inspiration from the remarkable image generation capabilities of current diffusion-based generative models~\cite{sohl2015deep,ho2020denoising,song2019generative}. 
The main inspiration of our new perspective lies in the score estimation part for both adversarial attack generation and image generation process within score based diffusion models~\cite{song2019generative}. We aim to answer a question \enquote{Can the score from a generative model be used to generate effective adversarial attacks?}
By answering this question, 
our objective is to develop a new victim model derived from image generation principles, obviating the need for specific victim model training, \eg~we seek to generate effective adversarial attacks for segmentation tasks without relying on a pre-trained segmentation model. Furthermore, 
given the absence of specific victim models, 
we are expecting effective adversarial attacks with good transferability that can be applied to models with different structures.

Our new perspective of generating victim model adversarial samples is related to attacks from
data distribution
\cite{zhu2021rethinking,Chen_2023_ICCV,li2022revisiting,lee2023robust,nie2022diffusion,wang2022guided,selvaraju2017grad}.
The core observation is that attack transferability is associated with the density of the ground truth distribution, with attacks aimed at low-density regions exhibiting better transferability.
In this case, the victim models are designed as task-aware, and gradients are obtained as pointing to the low-density region of the ground truth distribution. Differently, we propose a victim-model-agnostic distribution based attack, where good transferability is achieved via formulating the gradient of task related loss with image generation scores from score based diffusion models.

We summarize our main contributions as: 
\begin{enumerate}[] 
\item We introduce a novel adversarial attack method capable of generating sample-dependent adversarial samples without reliance on task specific victim model; 
\item We derived and demonstrated that effective adversarial samples can be generated solely based on the predictions of a generative model;
\item The image-dependent adversarial samples generated by our method can achieve effective adversarial attack with good transferability; 
\item Our method seamlessly integrates with a minimal number of queries and produces adversarial samples with performance comparable to query-based attacks.
\end{enumerate}

\section{Related Work}
\label{sec:Related_Work}
\subsection{Adversarial attacks for segmentation:} Adversarial attack tries to generate invisible perturbations
to evaluate model robustness, where the perturbation generation process can be defined as an optimization task. Here, we denote the classification loss of a neural network as $\mathcal{L}$, the neural network parameters as $\theta$, input data as $x$ and its corresponding one-hot encoded label as $y$. The objective of adversarial attacks~\cite{szegedy2013intriguing, adversarial_training, madry2017towards, Towards_Evaluating_Robustness_Neural_Networks} is to maximize the classification loss $\mathcal{L}(f_\theta(x^{\text{adv}}),y)$, such that the classifier will make wrong predictions for the attacked image $x^{\text{adv}}$.

Attacks on the segmentation tasks aim to make the model produce false predictions at the pixel level~\cite{ozbulak2019impact,treu2021fashion}.
Arnab \etal~\cite{arnab2018robustness} pioneered the assessment of adversarial samples in white-box attacks on a semantic segmentation model.
Additionally, Xie \etal~\cite{xie2017adversarial} highlighted the challenge of transferring white-box adversarial samples to models with different network structures in dense prediction tasks.
Arnab \etal~\cite{Metzen_2017_ICCV} introduced a generalized perturbation on semantic segmentation tasks, capable of causing a category in the prediction to vanish or yield a predefined output.
For more efficient white-box attacks on semantic segmentation tasks, Gu \etal~\cite{gu2022segpgd} increased the weights of pixels unsuccessfully attacked during gradient computation, while decreasing the weights of pixels successfully attacked.
In the area of black-box attacks, Zhang \etal~\cite{zhang2021data} proposed a universal adversarial perturbation for semantic segmentation based on the checkerboard format, which could attack all models without query. Li \etal~\cite{li2022query} reduced the number of queries for black-box attacks on the medical image segmentation task by learning the distribution of the impact of square perturbations on victim model attacks.

Among all those attacks for segmentation, the victim segmentation models are assumed available. In this paper, we rethink adversarial attack 
by adopting an image generation perspective. We introduce a new victim model without relying on any pre-existing segmentation models, leading to a distinct angle for the generation of attack samples.

\subsection{Black-box attacks:}  As the parameters/structures of the victim models are not known, black-box attacks
 rely on the predictions of the victim model or surrogate model to generate adversarial perturbations~\cite{maho2021surfree,ma2021simulating,tashiro2020diversity,rahmati2020geoda,feng2022boosting}. We roughly categorize the existing black-box attacks to
 query-based attacks~\cite{chen2017zoo,ilyas2018black,al2019sign,li2022query,andriushchenko2020square,liang2021parallel} and transfer-based attacks~\cite{qin2022boosting,long2022frequency,wu2020skip}. 

\textit{Query based} black-box attacks rely on querying the victim model to gradually generate adversarial perturbation, making the number of queries an important factor for effective attacks, where a large number of queries is favored for better performance. However, it may cause less efficient attack generation. 
To reduce the number of queries, Guo \etal~\cite{guo2019simple} constructed an orthogonal search space in the frequency domain.
\cite{andriushchenko2020square} added perturbations locally by randomly selecting a patch, and adding valid perturbations for iteration. 
On this basis,~\cite{li2022query} modeled the relationship between the effective perturbation and the ground truth, and learned the location distribution for more effective perturbation.

\textit{Transfer-based}  black-box attacks generate adversarial samples on the local white-box surrogate model and transfer the attacking properties of these samples to the victim model~\cite{qin2022boosting,long2022frequency,wu2020skip}. However, due to the gap between the white-box surrogate model and the black-box victim model, the transfer-based attacks are at the risk of overfitting the surrogate model.
In order to mitigate the overfitting issue,
\cite{zhou2018transferable,ganeshan2019fda,wang2021feature,wu2020boosting} proposed enhancing transferability in attacks through intermediate feature attack.
With the same goal, \cite{lin2020nesterov,wang2021enhancing,xiong2022stochastic,zhang2023transferable} adjusted the gradient calculation process of the surrogate model to avoid the overfitting issue. Zhu \etal~\cite{zhu2022toward} fine-tuned the surrogate models to match the classification probability gradient to the conditional probability gradient of data distributions. 
Additionally, \cite{wang2021admix,dong2019evading,wu2021improving} worked on improving the diversity of the samples to enhance the generalization ability of attacks.

\subsection{Data distribution driven attacks:}
Considering the optimization nature of attack generation, some recent work generates attacks via data distribution analysis.
Zhu \etal ~\cite{zhu2021rethinking} presented a transfer attack method that aligns the data distribution's gradient descent direction with the loss gradient descent direction.
Li \etal ~\cite{li2022revisiting} explained the effectiveness of the gradient-based attack method from the perspective of data distribution in graph neural networks.
Diffusion has also been used in attack tasks due to its powerful generative capabilities. Chen \etal ~\cite{Chen_2023_ICCV} introduced the diffusion denoising process into PGD-based~\cite{madry2017towards} white-box attack, reducing the perceptibility of adversarial samples. Xue \etal ~\cite{xue2024diffusion} utilized a diffusion model to bring each step of the adversarial sample closer to the clean distribution. Chen \etal \cite{chen2023diffusion} introduced stable diffusion to generate perturbations in latent feature space based on text prompts.
Differently, for the first time, we derive a new victim model from the correlation of diffusion-based model predictions with score, getting rid of the requirement of a segmentation model's existence, leading to a new generative perspective for attack generation.

\section{Our Method}
In this section, we present our image generation perspective to adversarial attack by first introducing preliminary knowledge on attacks in \cref{subsec_background}, which serves as the foundation for our way of generating attacks.
Then we present our new victim model in \cref{subsec_css_estimation}, which can be derived via image generation score in \cref{victim_attack_method}. We show the training process and model structure in \cref{subsec_network}.

\subsection{Background}
\label{subsec_background}
\subsubsection{White-box attacks}
The white-box attack methods necessitate a gradient-accessible victim model, as illustrated in \cref{Fig.attack_diff} (a). This type of attack method computes a loss function $\mathcal{L}$ based on model predictions $f_v(x)$ for a given sample $x$, where $v$ represents victim model parameters. Adversarial samples are then generated along the direction of increasing the loss term (FGSM~\cite{adversarial_training}):
\begin{equation}
\begin{aligned}
    \label{white_box}
x^{\text{adv}}=x+\epsilon \text{sign}(\nabla_{x}\mathcal{L}(f_v(x),y),
 \end{aligned}
\end{equation}
where $y$ is the ground truth of $x$,
$\epsilon $ is the perturbation rate, $\text{sign}$ is the sign function, and $\delta^{\text{adv}}=\epsilon \text{sign}(\nabla_{x}\mathcal{L}(f_v(x),y)$ is the adversarial perturbation, which is usually $\ell_p$ norm bounded to achieve perceptually invisible attacks, \ie~$\ell_p=\ell_\infty$.
FGSM~\cite{adversarial_training} achieves reasonable attacks for most simple scenarios. However, its single-step scheme limits its application for complex images.
PGD~\cite{madry2017towards} is then introduced as a multi-step variant:
\begin{equation}
\label{eq_pgd_attack}
    x^{\text{adv}}_{m+1} = \text{CLIP}_{x_0}\left(x^{\text{adv}}_m + \mu\text{sign} \left( \nabla_{x^{\text{adv}}_m} \mathcal{L}(f_v(x^{\text{adv}}_m), y) \right)\right),
\end{equation}
where $\mu$ is the step size, {$\text{CLIP}_{x_0}$ is a clip function around $x_0$, defined by perturbation rate $\epsilon$, and $m$ indexes the step.

\subsubsection{Query-based black-box attacks}
The query-based black-box attacks target the gradient-inaccessible victim model, as shown in \cref{Fig.attack_diff} (b). These methods maximize the loss function $\mathcal{L}$ of the model's prediction $f_v(x^{\text{adv}})$ by searching for perturbation  given the maximum number of queries:

\begin{equation}
\label{eq:black_box_query}
\begin{aligned}
    &\delta^{\text{adv}} = \argmax_{||\delta||_{p} \leq \delta_{d}} \mathcal{L}(f_v(x+\delta),y),
\end{aligned}
\end{equation}
where $\delta_{d}$ is the upper bound of perturbation, which can be defined as $\epsilon$, and $x$ is updated every each query: $x = x + \delta^{\text{adv}}$.
Query-based attacks repeat the above process until a stopping criterion is met, \eg~a certain number of queries or a desired level of adversarial perturbation is achieved.

\subsubsection{Transfer-based black-box attacks}
The transfer-based black-box attacks involve locally training a surrogate model $f_s$, followed by a white-box attack on this surrogate model as in \cref{white_box} or \cref{eq_pgd_attack}. The generated adversarial samples $x^{\text{adv}}$ are then transferred to the gradient-inaccessible victim model, as shown in \cref{Fig.attack_diff} (c). 
The primary challenge of transfer-based black-box attacks is how to avoid adversarial samples from falling into a local optimum at the surrogate model, so as to effectively attack the victim network.

\subsection{Conditional Segmentation Score Estimation}
\label{subsec_css_estimation}
To achieve effective attacks on segmentation, the primary objective is to obtain the optimized perturbation $\delta^{\text{adv}}$ by pushing the model to produce an inaccurate prediction for the attacked image $x^{\text{adv}}=x+\delta^{\text{adv}}$.
Considering a conditional distribution $p(y|x) = {p(x,y)}/{p(x)}$, where $x$ is an RGB image  and $y$ is the corresponding ground truth map, which is the accurate  task-related segmentation map, the objective of an effective attack is then to minimize the likelihood of $p(y|x^{\text{adv}})$, or the corresponding log-likelihood $\log p(y|x^{\text{adv}})$.

With Bayes' rule, we have:
\begin{equation}
    \label{attack_likelihood}
    \log p(y|x^{\text{adv}})=\log{\frac{p(x^{\text{adv}},y)}{p(x^{\text{adv}})}}
    =\log p(x^{\text{adv}},y) - \log p(x^{\text{adv}}),
\end{equation}

The optimal perturbation is then obtained by minimizing $\log p(y|x^{\text{adv}})$ namely $\delta^{\text{adv}}$, leading to:
\begin{equation}
\begin{aligned}
    \delta^{\text{adv}} &= \argmin_{||\delta||_{p} \leq \delta_{d}} \log p(y|x+\delta)\\
    &=\argmin_{||\delta||_{p} \leq \delta_{d}} \left(\log p(x+\delta,y) - \log p(x+\delta)\right),
\end{aligned}
\end{equation}
which is also equivalent to finding direction towards $-\nabla_{\delta^{\text{adv}}}\log p(y|x+\delta^{\text{adv}})$.
We then define three types of score, namely the conditional segmentation score $s_\theta(y|x^{\text{adv}})$, the conditional and unconditional image generation score $s_\theta(x^{\text{adv}}|y)$ and $s_\theta(x^{\text{adv}})$, respectively:
\begin{equation}
\left\{
    \begin{array}{lr}
         s_\theta(y|x^{\text{adv}}) = \nabla_{\delta^{\text{adv}}}\log p(y|x+\delta^{\text{adv}}),&  \\
         s_\theta(x^{\text{adv}}|y) = \nabla_{\delta^{\text{adv}}} \log p(x+\delta^{\text{adv}},y),&  \\
         s_\theta(x^{\text{adv}}) = \nabla_{\delta^{\text{adv}}} \log p(x+\delta^{\text{adv}}). &
    \end{array}
\right.
\end{equation}

Recalling the step-wise perturbation in \cref{eq_pgd_attack}, \ie $\mu\text{sign}\left( \nabla_{x^{\text{adv}}_m} \mathcal{L}(f_v(x^{\text{adv}}_m, y)) \right)$,  it relies on a segmentation victim model. Now, the gradient term is replaced by $-s_\theta(y|x^{\text{adv}})$, allowing the step-wise perturbation to be obtained through the conditional segmentation score.  Therefore, we have: 
\begin{equation}
\label{eq_conditional_seg_score_derivation}
   s_\theta(y|x^{\text{adv}}) =  s_\theta(x^{\text{adv}}|y) - s_\theta(x^{\text{adv}}),
\end{equation}
indicating the conditional segmentation score can be computed without a victim segmentation model, and the conditional and unconditional image generation score can be used instead to generate the adversarial attack from an image generation perspective.
 \cref{eq_conditional_seg_score_derivation} suggests that by computing both the conditional and unconditional image generation score, we can obtain the conditional segmentation score, sharing a similar spirit as CFG~\cite{ho2022classifier}.

The basic assumption of  \cref{eq_conditional_seg_score_derivation} is that the score perfectly match the gradient of the log data distribution~\cite{song2019generative}, which can be biased due to the limited training data. Following~\cite{ho2022classifier}, we define the final conditional segmentation score as weighted linear combination of the conditional and unconditional scores for image generation, leading to:
\begin{equation}
\label{eq_conditional_seg_score_derivation_weighted}
   s(y|x^{\text{adv}}) =  \omega\left(s_\theta(x^{\text{adv}}|y) - s_\theta(x^{\text{adv}})\right),
\end{equation}
where the hyper-parameter $\omega$ is used to cancel out the gap between the actual score and the estimated score. In summary, we replace the traditional transfer attack's process of estimating the loss gradient of a specific surrogate model with estimating the gradient of the data distribution density. Based on \cref{eq_conditional_seg_score_derivation_weighted}, our next step is computing scores for image generation, where we refer to score based diffusion models~\cite{sohl2015deep,ho2020denoising,song2019generative} due to its advance in mode coverage, leading to better distribution modeling compared with other likelihood based generative models~\cite{vae_raw,rezende2015variational}.

\noindent\textbf{Estimation of condition/unconditional scores for image generation with diffusion models:}
Given a series of non-negative incremental noises schedule $\{\alpha_{t}\} _{t=0}^{T}$,
we let $x_t$ denote the perturbed image based on $x_{t-1}$. With Gaussian transition kernel, $x_t$ is obtained via~\cite{ho2020denoising}:
\begin{equation}
\begin{aligned}
    \label{xt_xt-1_1}
 x_t = \sqrt{\alpha_{t}}x_{t-1} + \sqrt{1-\alpha_{t}}z_{t-1},
 \end{aligned}
\end{equation}
which shows the diffusion process that starts from $x_0$, namely the clean image or the initial state, and gradually destroys the image to obtain a noise $x_T$ that follows a standard normal distribution.
$z_{t-1}\sim \mathcal{N}(\textbf{0},\textbf{I})$ denotes the Gaussian noise. Accordingly, the generation process starts from the random noise $x_T\sim \mathcal{N}(\textbf{0},\textbf{I})$ to gradually remove the noise and recover the clean image $x_0$.

Diffusion models are then trained as the noise estimator $\gamma_\theta$ (or the negative score, namely $-s_\theta$). Similarly, a conditional diffusion model~\cite{song2020denoising} is trained to obtain the conditional score. With a task-related conditional diffusion model for 2D image generation,  we have access to both the conditional score $s_\theta(x|y)$ and unconditional score $s_\theta(x)$. In practice, with the conditional diffusion model $s_\theta(x|y)$, the unconditional diffusion model is obtained via $s_\theta(x)=s_\theta(x|y=\varnothing)$ following CFG~\cite{ho2022classifier}.

\subsection{Generating Attacks via Score Estimation}
\label{victim_attack_method}
With the proposed formulation to obtain conditional segmentation score via a weighted linear combination of conditional and unconditional scores for image generation (see  \cref{eq_conditional_seg_score_derivation_weighted}), we obtain the step-wise gradient term in  \cref{attack_likelihood}, by replacing $\nabla_{x^{\text{adv}}_m} \mathcal{L}(f_v(x^{\text{adv}}_m), y)$ in \cref{eq_pgd_attack}, we achieved freedom from attacks that require a specific victim model.
We now present our attacks generation pipeline.

Particularly, we initialize the adversarial perturbation $\delta^{\text{adv}}=0$, and define the step-wise perturbation before the sign activation function as:
\begin{equation}
\begin{aligned}
    \label{eq_step_wise_perturbation}
    \delta_m &=  - \sqrt{1-\alpha_{m}} \nabla_{x^{\text{adv}}_m}\log p(y|x^{\text{adv}}_m) \\  
    &= -\sqrt{1-\alpha_{m}} s(y|x^{\text{adv}}_m)\\
    &= -\sqrt{1-\alpha_{m}}\omega\left(s_\theta(x^{\text{adv}}_m|y) - s_\theta(x^{\text{adv}}_m)\right),
 \end{aligned}
\end{equation}
where we have $x^{\text{adv}}_{0} = x$, indicating the clean image.  
We follow  \cref{xt_xt-1_1} to add  noise to \textit{pseudo adversarial sample} $x^{\text{adv}}_m$ to diffuse it to random noise of standard normal distribution:
\begin{equation}
\begin{aligned}
    \label{adv_1add_noise}
 x^{{\text{adv}}}_{m+1} = & \text{CLIP}_{x} \left ( \sqrt{\alpha_{m}}\cdot x^{{\text{adv}}}_m+\sqrt{1-\alpha_{m}}\cdot \delta_m \right ), 
\end{aligned}
\end{equation}
where $\alpha_{m}$ denotes the noise schedule as in  \cref{xt_xt-1_1}. \cref{adv_1add_noise} provides the next step input for \cref{eq_step_wise_perturbation}.  
We repeat  \cref{eq_step_wise_perturbation} and  \cref{adv_1add_noise} to generate a sequence of the step-wise perturbation, and the proposed attack process gradually accumulates perturbation to generate our adversarial attack $\delta^{\text{adv}}$ via:
\begin{equation}
    \label{add_noise}
    \delta^{\text{adv}} =\text{CLIP}_{\epsilon} \left ( \mu \text{sign}(\delta_{m}) + \delta^{\text{adv}}\right ).
\end{equation}
The adversarial sample is obtained:
\begin{equation}
\begin{aligned}
    \label{adv_1final}
 x^{\text{adv}} = & \text{CLIP}_{x}  \left ( x + \delta^{\text{adv}} \right ). 
\end{aligned}
\end{equation}
It is worth noting that the final adversarial sample $x^{\text{adv}}$ is obtained from the clean sample $x$ and the accumulated perturbation $\delta^{\text{adv}}$, and is unrelated to the \textit{pseudo adversarial sample}  $x^{\text{adv}}_m$. $x^{\text{adv}}_m$ is only used to calculate the perturbation at each step. 

The overall attack algorithm is shown in \cref{xgt_alg}. In contrast to victim segmentation model aware white-box or black-box attacks, we bring a new perspective to the generation of adversarial attack from an image generation perspective. Further, different from the transfer-based methods, our method can combine with queries, enabling the selection of an optimal number of attack steps to
enhance the overall effectiveness of the attack.
\begin{algorithm}[t]  
\caption{
Generating Attacks via Score Estimation. 
}
\textbf{Input}:  Image $x$ and corresponding ground truth ${y}$, maximum number of queries $m_\text{max}$, noise estimator $-s_\theta$.\\
\textbf{Optional input:} The query loss $\mathcal{L}_{Q}$, the victim model $f_v$, and the threshold for query loss $\mathcal{L}_{best} =0$.\\
\textbf{Output}:  Attack sample $x^{\text{adv}}$
\begin{algorithmic}[1]
    \STATE  Initialize $x^{{\text{adv}}}_0 = x$, $ \delta^{\text{adv}}=0$.
    \FOR {$m \leftarrow 0$ to $m_\text{max}$}  
    \STATE  Compute step-wise perturbation $ \delta_{m}$ with  \cref{eq_step_wise_perturbation};
    \STATE  Generate pseudo adversarial sample $x^{\text{adv}}_{m+1}$ for the next step's input with \cref{adv_1add_noise};
    \STATE  Cumulative the adversarial attack $\delta^{\text{adv}}$ with  \cref{add_noise};
    \IF {\textbf{Query Victim Model:}}
        \STATE  Construct query sample  $x^{Q} = x^{{\text{adv}}}_m$;
        \IF{$\mathcal{L_{Q}}(f_v(x^{Q} ),y)>\mathcal{L}_{best}$}
            \STATE Update the optimal adversarial perturbation $\delta^{best}$ with $\delta^{\text{adv}}$: $\delta^{best} = \delta^{\text{adv}}$;
            \STATE Update loss threshold $\mathcal{L}_{best} = \mathcal{L_{Q}}(f_v(x^{Q}),y)$;
        \ENDIF
    \ENDIF
    \ENDFOR  
  \IF {\textbf{Query Target Model:}}
  \STATE  $\delta^{\text{adv}}=\delta^{best}$
  \ENDIF
\STATE Obtain final adversarial sample $x^{\text{adv}}$ with  \cref{adv_1final}.
\end{algorithmic}
\label{xgt_alg}
\end{algorithm}
\subsection{Network}
\label{subsec_network}
In the training stage, a conditional diffusion model $s_\theta(x|y)$ on the segmentation task is employed to estimate both the conditional and unconditional scores.
We designed a UNet structured~\cite{song2020denoising} conditional diffusion model $s_\theta(x,c)$ with the conditional variable $c$, where $s_\theta(x,c)$ is the same as $s_\theta(x|y)$ except that $c$ is designed to train the model stochastically, allowing both conditional and unconditional score estimation.
Particularly, we define the stochastic conditional variable $c$ as:
\begin{equation}
\label{pro_choose}
c =
    \begin{cases}
        y & \quad \mathrm{if} \quad \beta\geq 0.1,\\
        \varnothing & \quad \mathrm{if} \quad \beta< 0.1,
    \end{cases}
\end{equation}
where $\beta$ is a random number in the rang of $[0,1]$.

We simultaneously train the conditional diffusion model and the unconditional diffusion model within a single network.
In the unconditional diffusion branch, starting with a clean sample $x_0\in\mathbb{R}^{{H\times}W\times3}$, we set the maximum time step $t$ as 1000,
and apply Gaussian noise $z\in\mathbb{R}^{{H\times}W\times3}$ to generate the noisy sample
$x_t$ via $x_t = \sqrt{\bar{\alpha}_{t}}x_{0} + \sqrt{1-\bar{\alpha}_{t}}z$.
We normalize the range of values of $y$ to $[-0.5,0.5]$, and set the conditional variable as $c = -1 \in\mathbb{R}^{{H\times}W\times1}$ for the unconditional training process (note that, an empty tensor from Pytorch can also be used as the conditional variable for the unconditional training process, and we observe no performance difference with our general setting).

For the unconditional setting, we define the objective function as:
\begin{equation}
\begin{aligned}
    \label{loss_uncondi}
\mathcal{L}_{\mathrm{uncondi}}=\mathbb{E}_{t,x_0,z_t}\|\sqrt{1-\bar{\alpha}_{t}}s_\theta(x_t,-1,t)+z_t\|^2,
\end{aligned}
\end{equation}
where $z_t$ is the specific random noise for $x_t$.
For the conditional diffusion branch, the conditional variable is defined as $c=y$,
and the objective function is:
\begin{equation}
\begin{aligned}
    \label{loss_condi}
\mathcal{L}_{\mathrm{condi}}=\mathbb{E}_{t,x_0,z}\|\sqrt{1-\bar{\alpha}_{t}}s_\theta(x_t,y,t)+z_t\|^2.
 \end{aligned}
\end{equation}
 
We train conditional and unconditional branches alternately according to the random indicator $\beta$ in
\cref{pro_choose}.

\section{Experiment}
\noindent\textbf{Tasks:} To validate the applicability of our method, we have conducted experiments on a binary segmentation task~(camouflaged object detection~\cite{fan2020camouflaged}, COD) and a multi-class segmentation task~(semantic segmentation). \\
\noindent\textbf{Dataset:} Black-box attacks in real-world scenarios only have access to the victim model predictions, with no knowledge of other information. To simultaneously verify the effectiveness of the attack method across different models and training datasets, more aligned with real-world scenarios, we differentiated the training data for the victim model and the local~(surrogate/DDE) model.
\textit{For COD task,} we employ the COD10K training dataset~\cite{fan2020camouflaged}~(4040 images) for training the victim models, and  NC4K~\cite{yunqiu_cod21}~(4121 images) for training the condition-based score estimation model and surrogate models.
We then evaluate the attack performance using the COD10K testing dataset, providing a more robust validation of the algorithms' attack effectiveness in the presence of data isolation. 
\textit{For the semantic segmentation task,} due to the lack of similar datasets, we split the PASCAL VOC 2012 (VOC)~\cite{everingham2010pascal} training set~(10,582 images) into two equally sized datasets to train
the surrogate model and the victim model, respectively. The attack performance is evaluated on the VOC validation set.\\
\noindent\textbf{Models:} We compare our image generation based victim model with the conventional segmentation victim models with various backbones. \textit{For COD task,} we select five popular network structures as encoder backbone networks:
ViT~\cite{ranftl2021vision}, PVTv2~\cite{wang2022pvt}, ResNet50~\cite{he2016deep}, Swin~\cite{liu2021swin}, and Vgg~\cite{vgg_network}.
To enhance efficiency in the camouflaged object detection task, we attach the same decoder structure as in~\cite{mao2021generative} to the above backbones, ensuring a finer optimization of predictions for all features extracted by the backbone network. 
\textit{For the semantic segmentation task,} we employed four backbones with representative networks: DeepLabV3+~\cite{deeplabv3} with MobileNet~\cite{mobilenets} backbone~(DL3Mob), DeepLabV3+ with ResNet101~\cite{he2016deep} backbone~(DL3R101), PSPNet~\cite{pspnet} with ResNet50~\cite{he2016deep} backbone~(PSPR50), and FCN~\cite{fcn} net with VGG16~\cite{vgg_network} backbone~(FCNV16).
All backbones are initialized with pre-trained model trained on ImageNet.

\begin{table*}[!pht]
  \centering
  \renewcommand{\arraystretch}{1.10}
  \renewcommand{\tabcolsep}{0.7mm}
  \caption{Performance comparison with the transfer-based black-box attack method on COD task using ViT as backbone of the surrogate model.}
 \resizebox{\linewidth}{!}{
 \begin{tabular}{l|cccc|cccc|cccc|cccc|cccc} 
 \bottomrule[0.75pt]
Victim  & \multicolumn{4}{c|}{ViT~\cite{ranftl2021vision}} & \multicolumn{4}{c|}{PVTv2~\cite{wang2022pvt}}& \multicolumn{4}{c|}{R50~\cite{he2016deep}}  &  \multicolumn{4}{c|}{Swin~\cite{liu2021swin}}  &  \multicolumn{4}{c}{Vgg~\cite{vgg_network}} \\   \hline
     &$\mathcal{M}\uparrow$ &$CC\downarrow$ & $S_{\alpha}\downarrow$&$E_{\xi}\downarrow$&$\mathcal{M}\uparrow$ &$CC\downarrow$ & $S_{\alpha}\downarrow$&$E_{\xi}\downarrow$&$\mathcal{M}\uparrow$ &$CC\downarrow$ & $S_{\alpha}\downarrow$&$E_{\xi}\downarrow$&$\mathcal{M}\uparrow$ &$CC\downarrow$ & $S_{\alpha}\downarrow$&$E_{\xi}\downarrow$&$\mathcal{M}\uparrow$ &$CC\downarrow$ & $S_{\alpha}\downarrow$&$E_{\xi}\downarrow$\\
  \hline
Baseline  & .026 & .781 & .852 & .924 & .024 & .797 & .866 & .932& .041 & .681 & .792 & .864 & .029 & .759 & .841 & .916 & .048 & .644 & .764 & .836\\

MI-FGSM~\cite{dong2018boosting}  & \textbf{.264} & \textbf{.287} & \textbf{.497} & \textbf{.491} & \textbf{.090} & .562 & .718 & .766 & .105 & .451 & .654 & .713 & .074 & .574 & .730 & .798 & .134 & .368 & .594 & .661 \\
$\text{DI}^{2}\text{-FGSM}$~\cite{xie2019improving}& .080 & .634 & .757 & .792 & .042 & .720 & .821 & .876 & .056 & .628 & .764 & .828 & .040 & .703 & .811 & .881 &  .064 & .592 & .737 & .804 \\
ILA~\cite{huang2019enhancing}& .121 & .480 & .654 & .701 & .043 & .704 & .808 & .874 & .064 & .580 & .732 & .804 & .043 & .687 & .796 & .876 & .084 & .504 & .682 & .753 \\
NAA~\cite{zhang2022improving}& .048 & .699 & .788 & .865 & .037 & .741 & .827 & .895 & .053 & .631 & .760 & .830
 & .038 & .719 & .812 & .888 & .074 & .540 & .701 & .773 \\  
RAP~\cite{qin2022boosting} & .068 & .536 & .691 & .802 &.046 & .665 & .777 & .868 & .066 & .544 & .705 & .794 & .047 & .645 & .765 & .862 & .083 & .468 & .658 & .749  \\ 

\textbf{Ours} & {.069} & {.570} &{ .708 }& {.802} & \textbf{.090} & \textbf{.497} & \textbf{.665} & \textbf{.756} &  \textbf{.123} & \textbf{.364} & \textbf{.596} & \textbf{.670} & \textbf{.088} & \textbf{.482} & \textbf{.658} & \textbf{.756} & \textbf{.160} & \textbf{.270} & \textbf{.526} & \textbf{.617}
\\ 
     \toprule[0.75pt]
     \end{tabular}
     }
  \label{tab:transfer_compare_dpt}
\end{table*}

\begin{table*}[!pht]
  \centering
  \renewcommand{\arraystretch}{1.10}
  \renewcommand{\tabcolsep}{0.7mm}
   \caption{Performance comparison with the transfer-based black-box attacks on COD task using ResNet50 as the surrogate model backbone.  \label{tab:transfer_compare_R50}}
 \resizebox{\linewidth}{!}{ 
 \begin{tabular}{l|cccc|cccc|cccc|cccc|cccc} 
 \bottomrule[0.75pt]
  Victim  & \multicolumn{4}{c|}{ViT }& \multicolumn{4}{c|}{PVTv2 }& \multicolumn{4}{c|}{R50 }  &  \multicolumn{4}{c|}{Swin }  &  \multicolumn{4}{c}{Vgg } \\ \hline
     &$\mathcal{M}\uparrow$ &$CC\downarrow$ & $S_{\alpha}\downarrow$&$E_{\xi}\downarrow$&$\mathcal{M}\uparrow$ &$CC\downarrow$ & $S_{\alpha}\downarrow$&$E_{\xi}\downarrow$&$\mathcal{M}\uparrow$ &$CC\downarrow$ & $S_{\alpha}\downarrow$&$E_{\xi}\downarrow$&$\mathcal{M}\uparrow$ &$CC\downarrow$ & $S_{\alpha}\downarrow$&$E_{\xi}\downarrow$&$\mathcal{M}\uparrow$ &$CC\downarrow$ & $S_{\alpha}\downarrow$&$E_{\xi}\downarrow$\\
  \hline
Baseline & .026 & .781 & .852 & .924 & .024 & .797 & .866 & .932& .041 & .681 & .792 & .864 & .029 & .759 & .841 & .916 & .048 & .644 & .764 & .836\\
MI-FGSM~\cite{dong2018boosting}  & .037 & .721 & .817 & .891 & .042 & .701 & .808 & .879 & .280 & .269 & .496 & .487 & .043 & .683 & .793 & .873 & .101 & .447 & .649 & .721  \\
$\text{DI}^{2}\text{-FGSM}$~\cite{xie2019improving}&  .035 & .732 & .825 & .895 & .045 & .708 & .814 & .874 & .312 & .322 & .503 & .485 & .039 & .701 & .808 & .885 & .115 & .466 & .659 & .702 \\
ILA~\cite{huang2019enhancing}  &  .035 & .739 & .822 & .898 & .043 & .715 & .811 & .884 &\textbf{.527} & \textbf{.071} & .\textbf{262 }& \textbf{.310} & .040 & .700 & .800 & .883 & .127 & .377 & .596 & .682 \\
IIA~\cite{zhu2021rethinking}  &  .035 & .721 & .814 & .893 & .040 & .693 & .803 & .879 & .173 & .237 & .513 & .603 &.044 & .668 & .786 & .870 & .161 & .292 & .541 & .631 \\
NAA~\cite{zhang2022improving}  & .046 & .684 & .785 & .863 & .057 & .646 & .764 & .837 & .286 & .109 & .394 & .447 & .048 & .659 & .772 & .855 & .130 & .341 & .574 & .645 \\  
RAP~\cite{qin2022boosting}   & .036 & .732 & .816 & .898 & .039 & .719 & .813 & .892 & .125 & .324 & .567 & .672 & .041 & .693 & .793 & .881 & .085 & .462 & .652 & .738 \\ 
\textbf{Ours} & \textbf{.069 }& \textbf{.570 }& \textbf{.708} & \textbf{.802} &\textbf{ .090} & \textbf{.497} &\textbf{ .665} & \textbf{.756} &  .123 & .364 & .596 & .670 & \textbf{.088} & \textbf{.482} & \textbf{.658} & \textbf{.756} & \textbf{.160} & \textbf{.270} & \textbf{.526} & \textbf{.617}
\\ 
     \toprule[0.75pt]
     \end{tabular}
     }
    
\end{table*}

\begin{table*}[!pht]
  \centering
  \small
  \renewcommand{\arraystretch}{1.10}
  \renewcommand{\tabcolsep}{0.0mm}
  \caption{Performance comparison with the transfer-based black-box attack method on semantic segmentation task using PSPR50 or DL3Mob as the surrogate model.}
 \resizebox{\linewidth}{!}{
 \begin{tabular}{l|cc|cc|cc|cc|cc|cc|cc|cc} 
 \bottomrule[1pt]
 Surrogate& \multicolumn{8}{c|}{PSPR50} & \multicolumn{8}{c}{DL3Mob}   \\ \hline
 Victim & \multicolumn{2}{c|}{PSPR50} & \multicolumn{2}{c|}{DL3R101}& \multicolumn{2}{c|}{DL3Mob}  &  \multicolumn{2}{c|}{FCNV16} & \multicolumn{2}{c|}{PSPR50} & \multicolumn{2}{c|}{DL3R101}& \multicolumn{2}{c|}{DL3Mob}  &  \multicolumn{2}{c}{FCNV16}  \\ \hline
     &${mIoU}\downarrow$ &$ACC\downarrow$ &${mIoU}\downarrow$ &$ACC\downarrow$ &${mIoU}\downarrow$ &$ACC\downarrow$ &${mIoU}\downarrow$ &$ACC\downarrow$ &${mIoU}\downarrow$ &$ACC\downarrow$ &${mIoU}\downarrow$ &$ACC\downarrow$ &${mIoU}\downarrow$ &$ACC\downarrow$ &${mIoU}\downarrow$ &$ACC\downarrow$ \\
  \hline
Baseline  & 0.716 & 0.920  & 0.707 & 0.916 & 0.642 & 0.897 &  0.258 & 0.785 & 0.716 & 0.920  & 0.707 & 0.916 & 0.642 & 0.897 &  0.258 & 0.785 \\
PGD~\cite{madry2017towards}  &  0.159  &  0.550   &  0.615  &  0.883  &  0.464  &  0.825  &  0.194  &  0.733 &  0.527  &  0.854  &  0.604  &  0.879  &  0.121  &  0.364  &  0.200  &  0.739\\
SegPGD~\cite{gu2022segpgd}    &  0.139  &  0.546   &  0.617  &  0.885 &  0.467  &  0.829   &  0.193  &  0.733 &  0.531  &  0.857  &  0.609  &  0.882  &  0.072  &  0.309   &  0.197  &  0.742 \\ \hline
MI-FGSM~\cite{dong2018boosting}   &  0.297  &  0.629 &  0.589  &  0.870 &  0.455  &  0.819  &  0.192  &  0.734 &  0.507  &  0.843  &  0.569  &  0.861  &  0.167  &  0.417 &  0.189  &  0.731\\
$\text{DI}^{2}\text{-FGSM}$~\cite{xie2019improving} &  0.356  &  0.665  &  0.495  &  0.819 &  0.500  &  0.831 &  0.215  &  0.738  &  0.564  &  0.860  &  0.609  &  0.873  &  0.236  &  0.509 &  0.218  &  0.742\\
ILA~\cite{huang2019enhancing}  &  0.151  &  0.616  &  0.407  &  0.782  &  0.341  &  0.753 &  0.171  & \textbf{ 0.685} &  0.456  &  0.817    &  0.513  &  \textbf{0.828}   &  0.164  &  0.527 &  0.175  &  \textbf{0.690} \\
NAA~\cite{zhang2022improving} &  \textbf{0.076}  &  \textbf{0.535}  &  \textbf{0.367}  &  \textbf{0.782}  &  \textbf{0.248} &  \textbf{0.736}  &  \textbf{0.145}  &  {0.712}  & \textbf{0.416}  &  \textbf{0.806}   &  0.502  &  0.837 &  \textbf{0.059}  &  \textbf{0.358}  &  \textbf{0.159}  &  0.716  \\  
RAP~\cite{qin2022boosting}     &  0.371  &  0.798  &  0.501  &  0.835 &  0.416  &  0.813  &  0.184  &  0.752 &  0.456  &  0.827  &  \textbf{0.501}  &  0.834  &  0.309  &  0.774 &  0.178  &  0.747 \\ 
IIA~\cite{zhu2021rethinking}   &  0.408  &  0.823 &  0.513  &  0.843  &  0.439  &  0.825  &  0.197  &  0.741 & - & -  & - & -  & - & -  & - & - \\ 
\textbf{Ours}     &  0.520  &  0.848    &  0.572  &  0.866 &  0.456  &  0.824   &  0.185  &  0.719 &  0.520  &  0.848    &  0.572  &  0.866 &  0.456  &  0.824   &  0.185  &  0.719\\   
     \toprule[1pt]
     \end{tabular}
     }
  \label{tab:transfer_compare_segment_psp}
\end{table*}

\noindent\textbf{Attack Setting:} Following conventional practice, we use
$\ell_{\infty}$ perturbation with a maximum allowable perturbation of $\delta_{d}=8/255$. In our method, the attack step size $\mu=2/255$. 
We set $m_{\text{max}}=30$ for both tasks on our transfer-based models.
In \cref{eq_conditional_seg_score_derivation_weighted}, $\omega$ is empirically set to 90 (model robustness analysis is performed to explain the sensitivity of our model~\wrt~$\omega$). The step size and iterations for other methods are configured according to the original papers. 
 For fair comparison, we refrain from employing ensemble settings in any methods. We solely compare the results of the attack algorithms independently proposed in all papers, avoiding comparisons involving the superimposition of other attack algorithms. \\
\noindent\textbf{Evaluation Metrics:} \textit{For COD task,}
correlation coefficient $CC$ is used to evaluate
attack performance, and
Mean Absolute Error ($\mathcal{M}$), Mean E-measure ($E_{\xi}$)~\cite{Fan2018Enhanced} and S-measure ($S_{\alpha}$)~\cite{fan2017structure} are adopted to
evaluate the segmentation accuracy.
\textit{For the semantic segmentation task,}
Mean Intersection over Union (mIoU) and Pixel Accuracy (ACC) are used to to evaluate the adversarial robustness. \\
\noindent\textbf{Training details:} We train diffusion model $s_\theta$ with structure from~\cite{song2020denoising} using
Pytorch, where both the encoder and decoder are initialized by default.
We resize all images with the ground truth image to $384 \times 384$ for the COD task and $480 \times 480$ for the semantic segmentation task. 
The maximum training step is 980,000 for COD and 230,000 for semantic segmentation. The learning rate is 2e-5 with Adam optimizer. The batch size is 12 for COD and 8 for semantic segmentation on four RTX 3090 GPUs.

\subsection{Performance comparison} 
\subsubsection{Quantitative comparison:}
We evaluate the effectiveness of our proposed attack method by comparing it with six transfer-based black-box attack methods.
Momentum Iterative Fast Gradient Sign Method~(MI-FGSM)~\cite{dong2018boosting} improved the transferability of white-box attack through a gradient momentum iterative method based on FGSM.
Diverse Inputs Iterative Fast Gradient Sign Method
($\text{DI}^{2}\text{-FGSM}$)~\cite{xie2019improving} employed a diverse input iterative fast gradient sign method to perform a white-box attack and transfer the adversarial samples to the victim model.
Intermediate Level Attack~(ILA)~\cite{huang2019enhancing} increased perturbation on the middle layer of the neural network based on pre-generated adversarial samples to enhance the transferability of attack.
Intrinsic Adversarial Attack~(IIA)~\cite{zhu2021rethinking} searched the parameters of the residual module to align the direction of data distribution decline with the rise in loss function value.
Neuron Attribution-based Attacks~(NAA)~\cite{zhang2022improving} conducted feature-level attacks by estimating neuronal importance.
Reverse Adversarial Perturbation~(RAP)~\cite{qin2022boosting} transformed the search optimal point of the attack into a search neighborhood optimal point to avoid perturbation into the local optimum.
The model trained with clean samples is denoted as the Baseline.

\begin{figure*}[!hpt]
   \centering
   \begin{tabular}{{c@{ } c@{ } c@{ } c@{ } c@{ }c@{ } c@{ } c@{ }c@{ } }}
 \rotatebox{90}{\small ~~~Image}& 
 {\includegraphics[width=0.13\linewidth,height=0.10\linewidth]{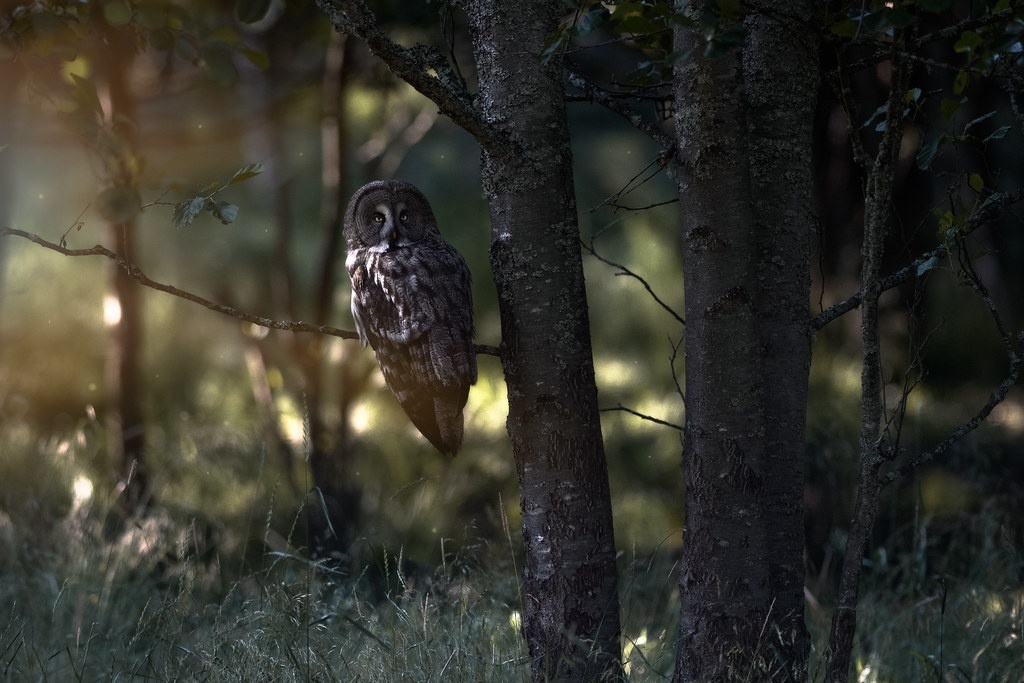}}&
 {\includegraphics[width=0.13\linewidth,height=0.10\linewidth]{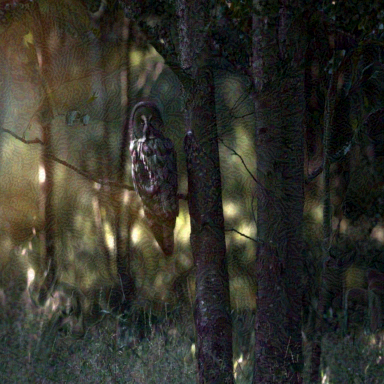}}&
 {\includegraphics[width=0.13\linewidth,height=0.10\linewidth]{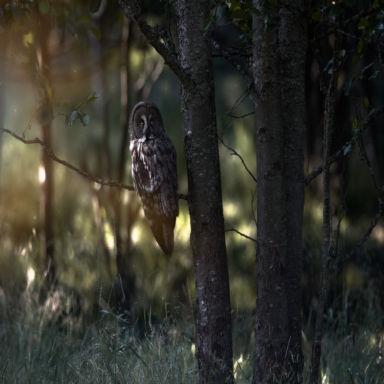}}&
 {\includegraphics[width=0.13\linewidth,height=0.10\linewidth]{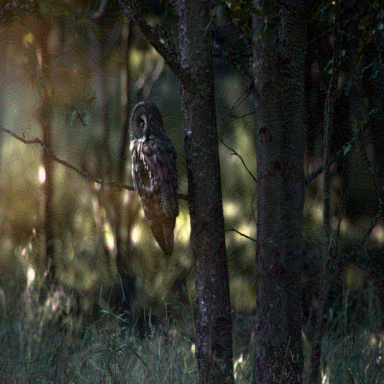}}&
 {\includegraphics[width=0.13\linewidth,height=0.10\linewidth]{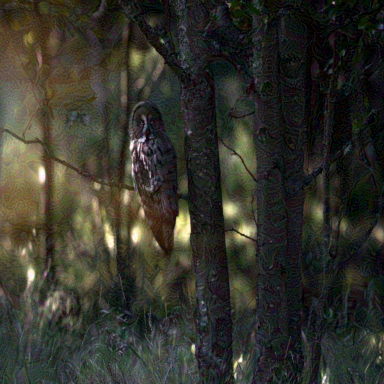}}&
 {\includegraphics[width=0.13\linewidth,height=0.10\linewidth]{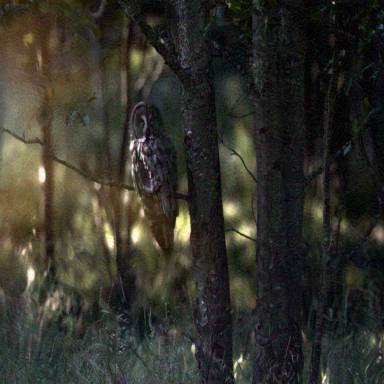}}&
 {\includegraphics[width=0.13\linewidth,height=0.10\linewidth]{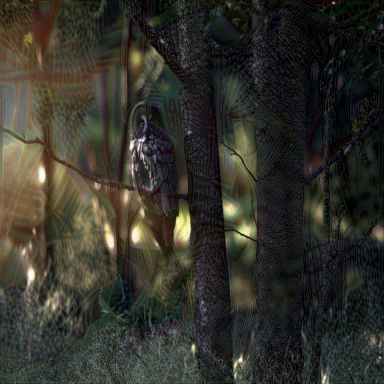}}
 \\
\rotatebox{90}{~~~~ \small R50}&  {\includegraphics[width=0.13\linewidth,height=0.10\linewidth]{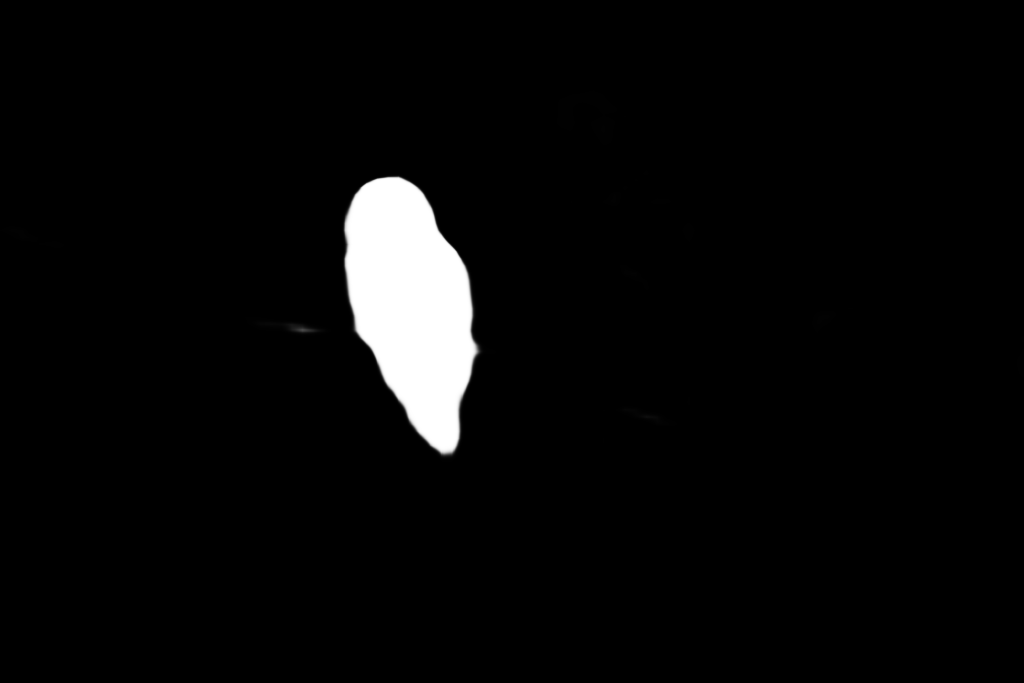}}&
 {\includegraphics[width=0.13\linewidth,height=0.10\linewidth]{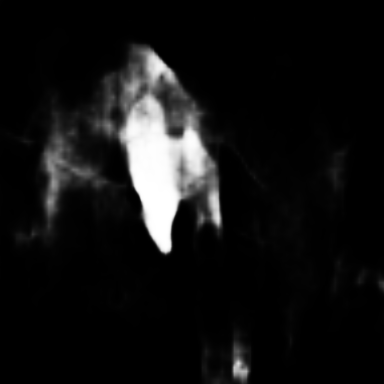}}&
 {\includegraphics[width=0.13\linewidth,height=0.10\linewidth]{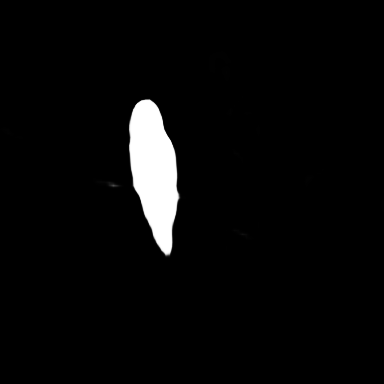}}&
 {\includegraphics[width=0.13\linewidth,height=0.10\linewidth]{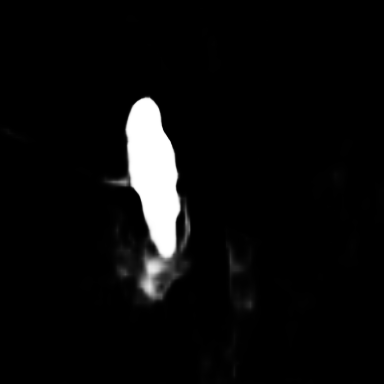}}&
 {\includegraphics[width=0.13\linewidth,height=0.10\linewidth]{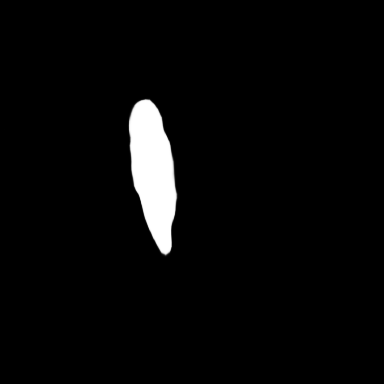}}&
 {\includegraphics[width=0.13\linewidth,height=0.10\linewidth]{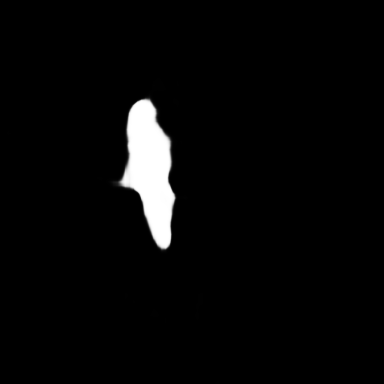}}&
 {\includegraphics[width=0.13\linewidth,height=0.10\linewidth]{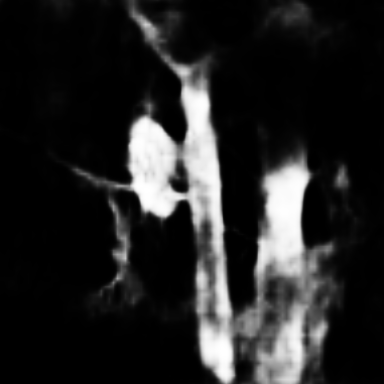}}
 \\
 \rotatebox{90}{~~~~\small Vgg}&  {\includegraphics[width=0.13\linewidth,height=0.10\linewidth]{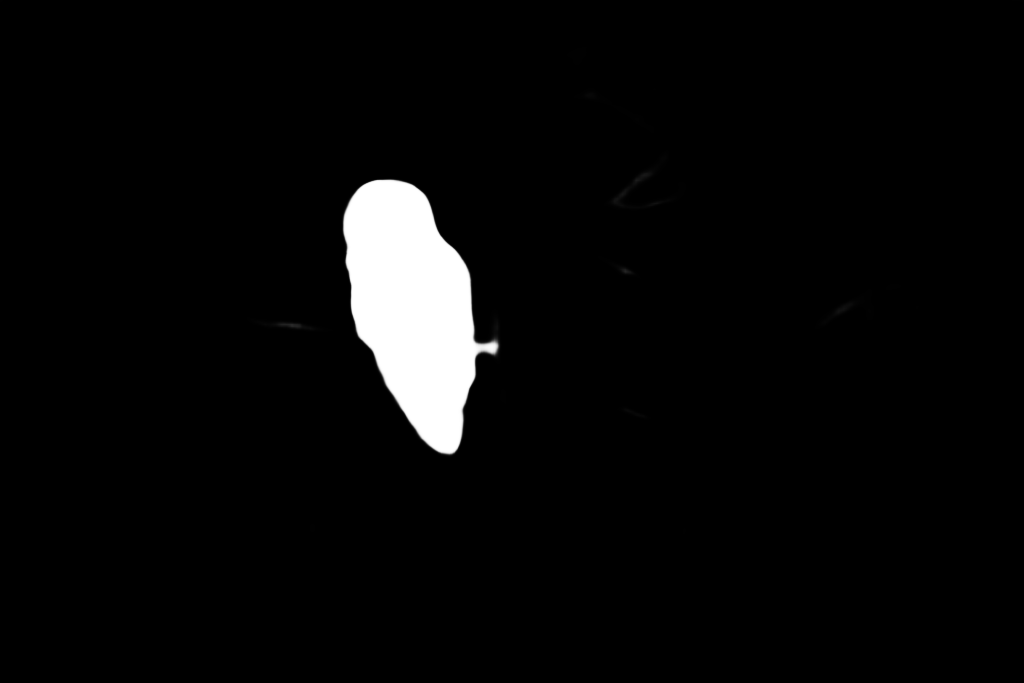}}&
 {\includegraphics[width=0.13\linewidth,height=0.10\linewidth]{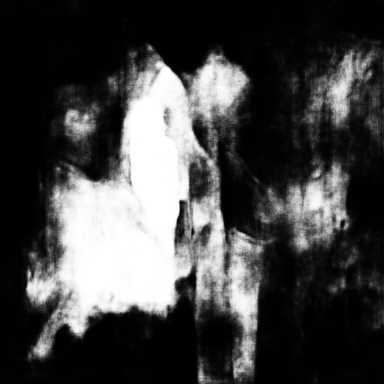}}&
 {\includegraphics[width=0.13\linewidth,height=0.10\linewidth]{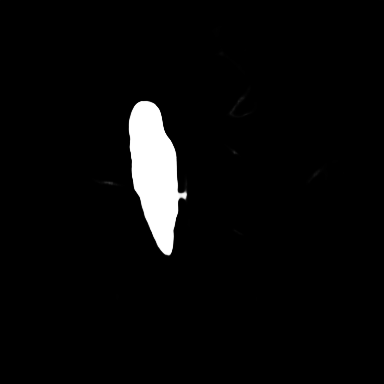}}&
 {\includegraphics[width=0.13\linewidth,height=0.10\linewidth]{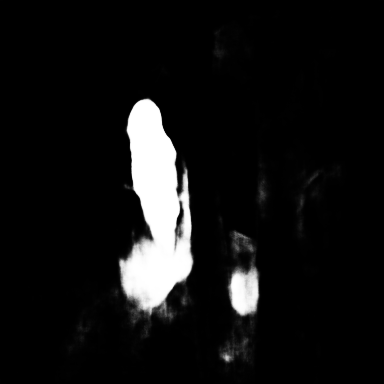}}&
 {\includegraphics[width=0.13\linewidth,height=0.10\linewidth]{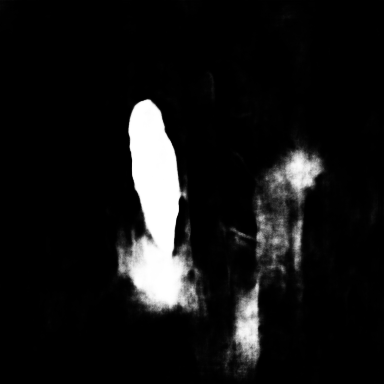}}&
 {\includegraphics[width=0.13\linewidth,height=0.10\linewidth]{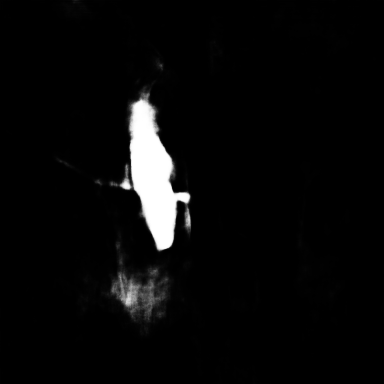}}&
 {\includegraphics[width=0.13\linewidth,height=0.10\linewidth]{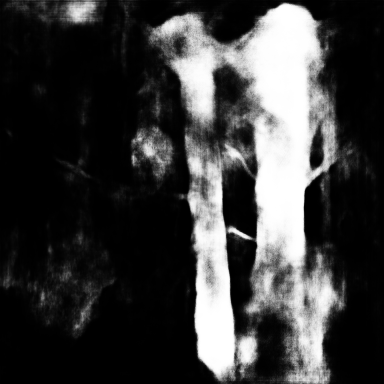}}
 \\
  \rotatebox{90}{~~~~\small PVTv2}&  {\includegraphics[width=0.13\linewidth,height=0.10\linewidth]{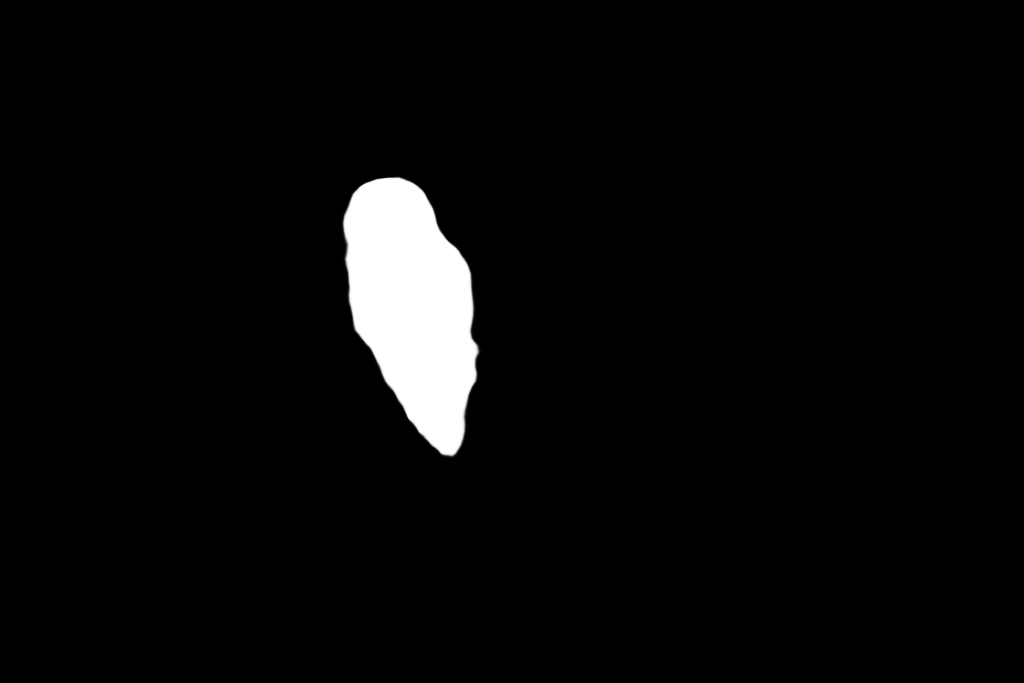}}&
 {\includegraphics[width=0.13\linewidth,height=0.10\linewidth]{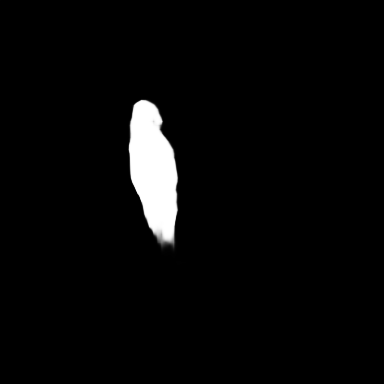}}&
 {\includegraphics[width=0.13\linewidth,height=0.10\linewidth]{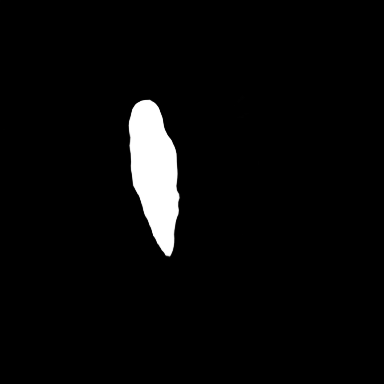}}&
 {\includegraphics[width=0.13\linewidth,height=0.10\linewidth]{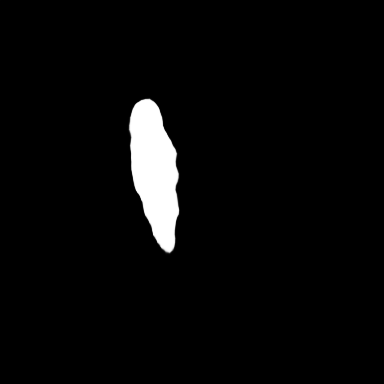}}&
 {\includegraphics[width=0.13\linewidth,height=0.10\linewidth]{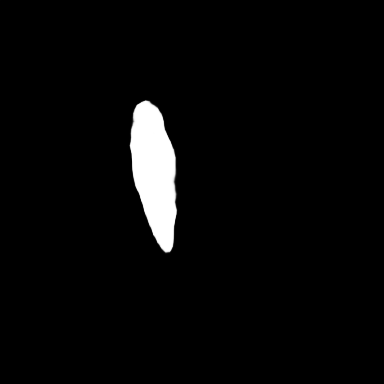}}&
 {\includegraphics[width=0.13\linewidth,height=0.10\linewidth]{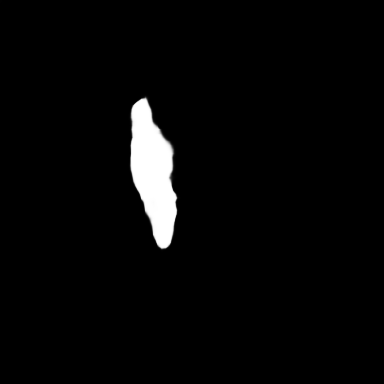}}&
 {\includegraphics[width=0.13\linewidth,height=0.10\linewidth]{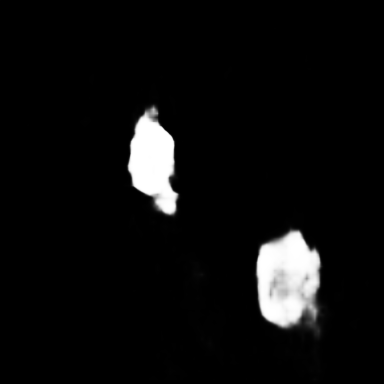}}
 \\
   & \small{Clean} 
   & {\small{MI-FGSM~\cite{dong2018boosting}}}  &{\small{$\text{DI}^{2}\text{-FGSM}$~\cite{xie2019improving}}}  & {\small{ILA~\cite{huang2019enhancing}}}  & {\small{NAA~\cite{zhang2022improving}}} & {\small{RAP~\cite{qin2022boosting}}} & {\small{Ours}}  \\
   \end{tabular} 
    \caption{Visual comparison of transfer-based black-box attacks on COD task with ViT as the surrogate model backbone. The first column represents the clean images and the corresponding predictions.} 
    \label{fig:transfer_comparison}
\end{figure*}

\begin{figure*}[!pht]
   \centering
   \begin{tabular}{{c@{ } c@{ } c@{ } c@{ } c@{ }c@{ } c@{ } c@{ }c@{ } }}
 \rotatebox{90}{ ~~ \small Image}& 
 {\includegraphics[width=0.13\linewidth,height=0.10\linewidth]{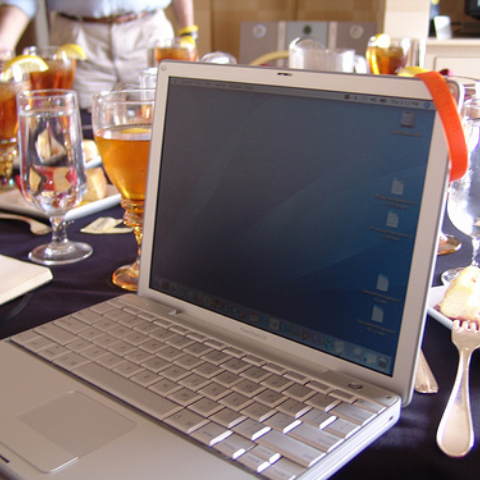}}&
 {\includegraphics[width=0.13\linewidth,height=0.10\linewidth]{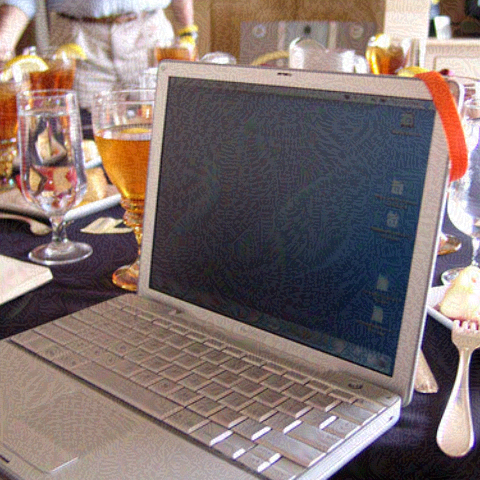}}&
 {\includegraphics[width=0.13\linewidth,height=0.10\linewidth]{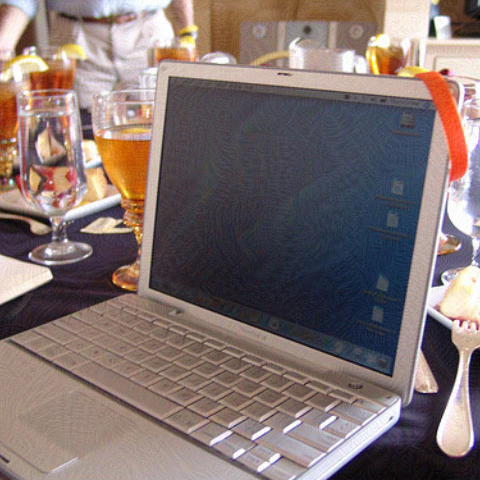}}&
 {\includegraphics[width=0.13\linewidth,height=0.10\linewidth]{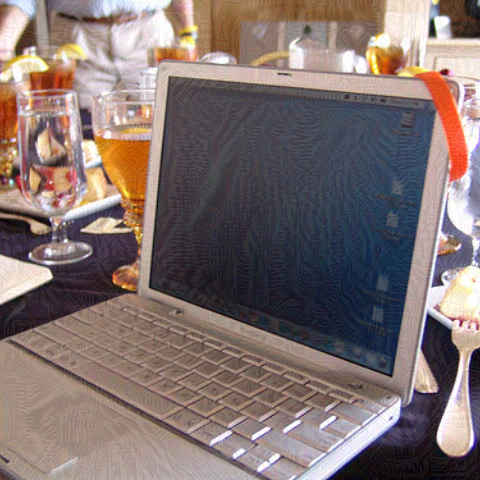}}&
 {\includegraphics[width=0.13\linewidth,height=0.10\linewidth]{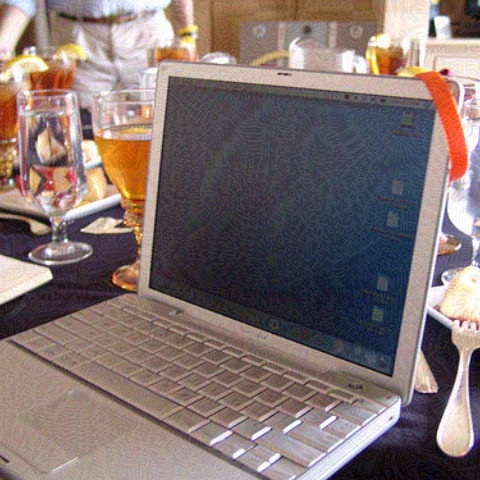}}&
 {\includegraphics[width=0.13\linewidth,height=0.10\linewidth]{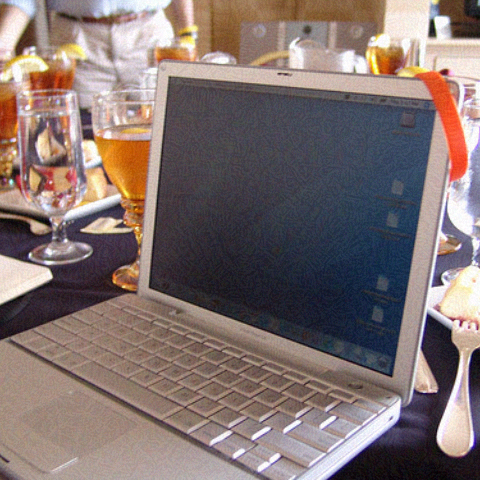}}&
 {\includegraphics[width=0.13\linewidth,height=0.10\linewidth]{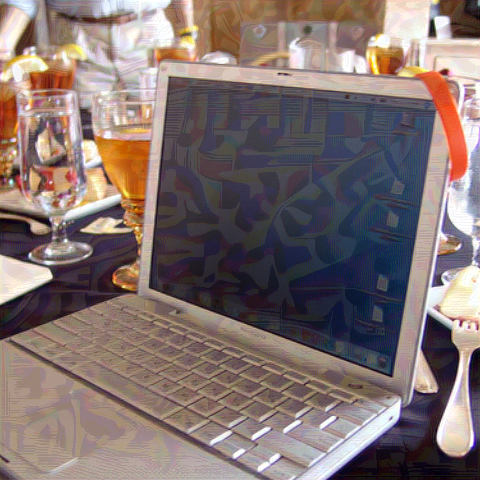}}
 \\
 \rotatebox{90}{~~ \small PSPR50}& 
 {\includegraphics[width=0.13\linewidth,height=0.10\linewidth]{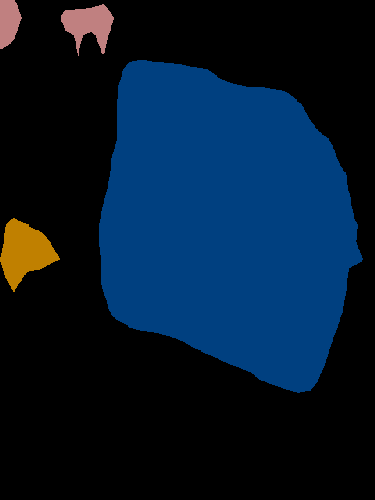}}&
 {\includegraphics[width=0.13\linewidth,height=0.10\linewidth]{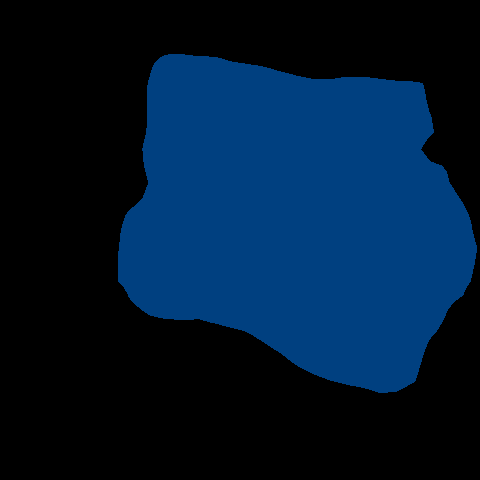}}&
 {\includegraphics[width=0.13\linewidth,height=0.10\linewidth]{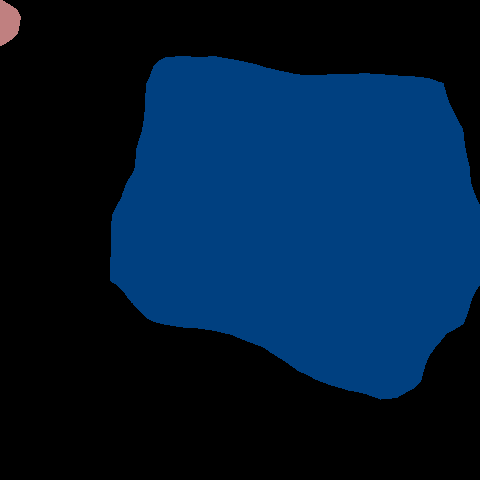}}&
 {\includegraphics[width=0.13\linewidth,height=0.10\linewidth]{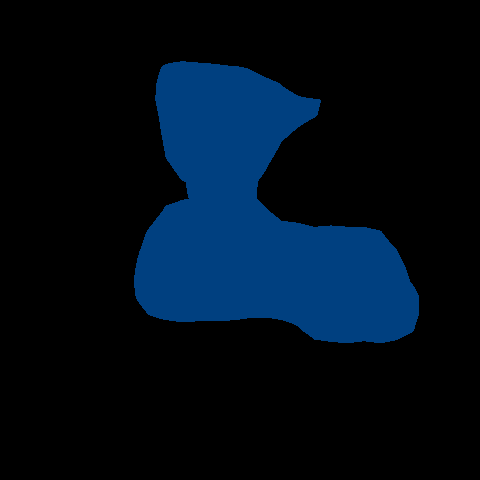}}&
 {\includegraphics[width=0.13\linewidth,height=0.10\linewidth]{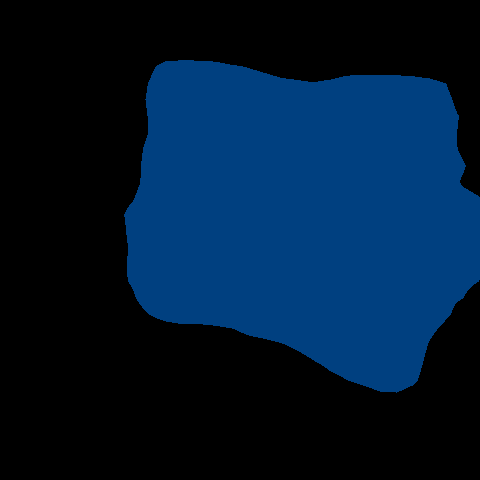}}&
 {\includegraphics[width=0.13\linewidth,height=0.10\linewidth]{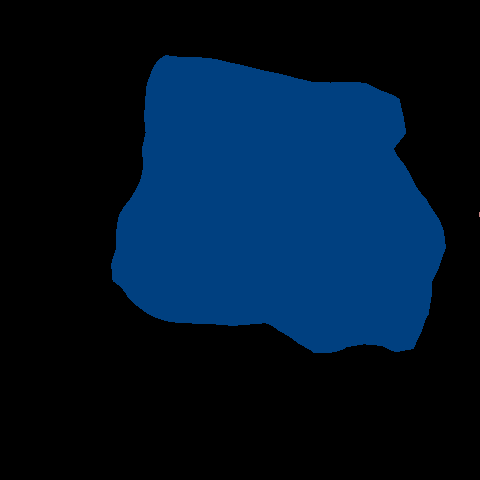}}&
 {\includegraphics[width=0.13\linewidth,height=0.10\linewidth]{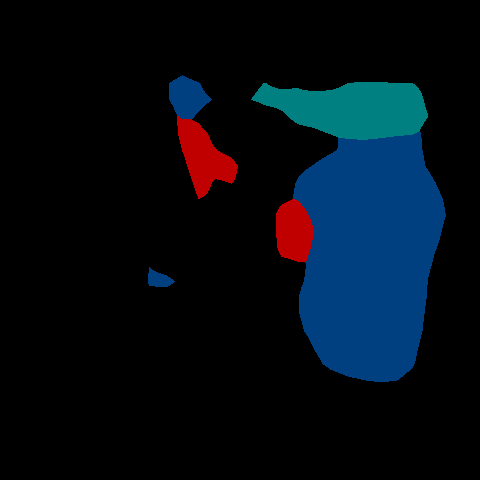}}
 \\
 \rotatebox{90}{~ \small DL3R101}& 
 {\includegraphics[width=0.13\linewidth,height=0.10\linewidth]{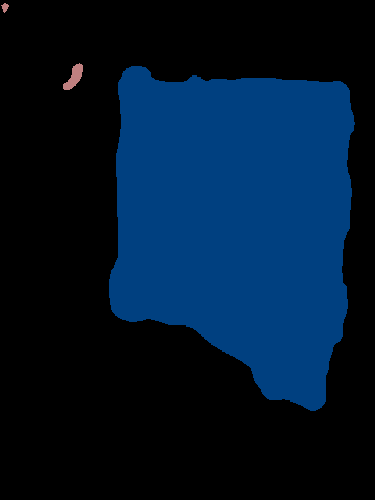}}&
 {\includegraphics[width=0.13\linewidth,height=0.10\linewidth]{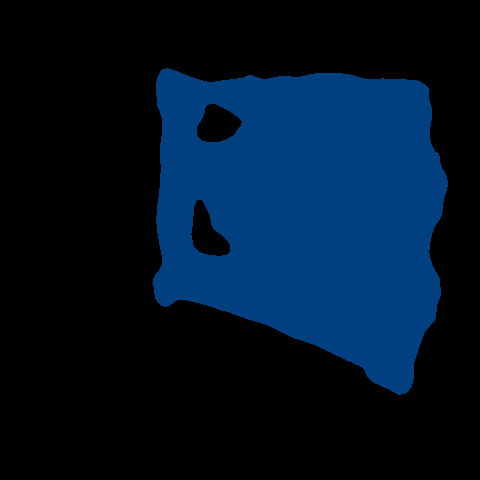}}&
 {\includegraphics[width=0.13\linewidth,height=0.10\linewidth]{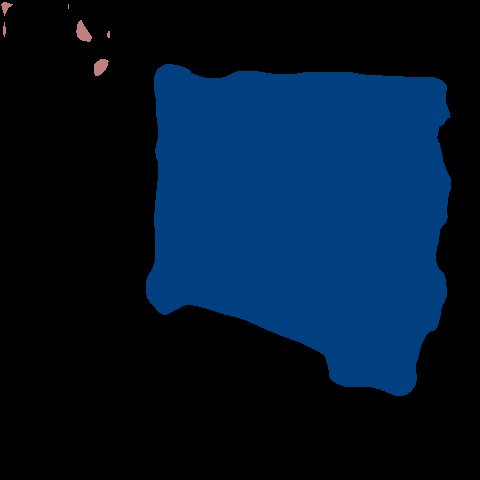}}&
 {\includegraphics[width=0.13\linewidth,height=0.10\linewidth]{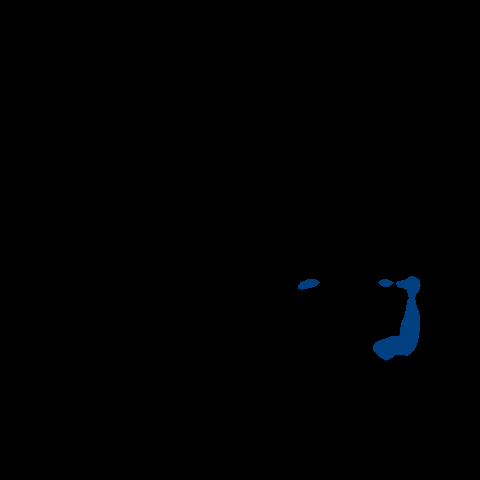}}&
 {\includegraphics[width=0.13\linewidth,height=0.10\linewidth]{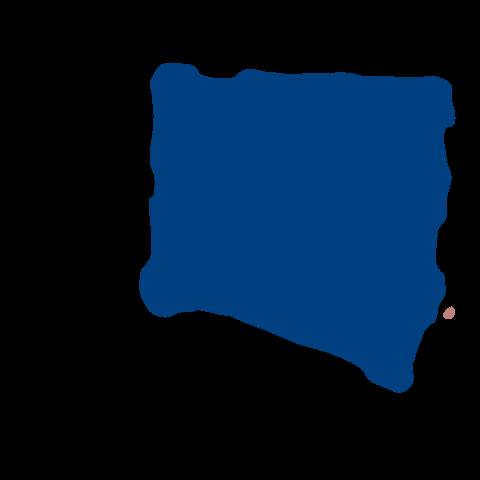}}&
 {\includegraphics[width=0.13\linewidth,height=0.10\linewidth]{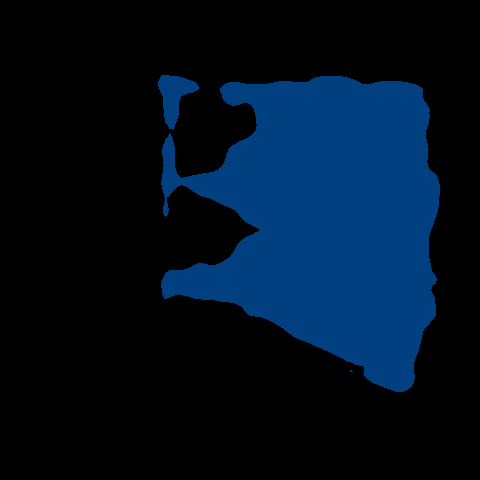}}&
 {\includegraphics[width=0.13\linewidth,height=0.10\linewidth]{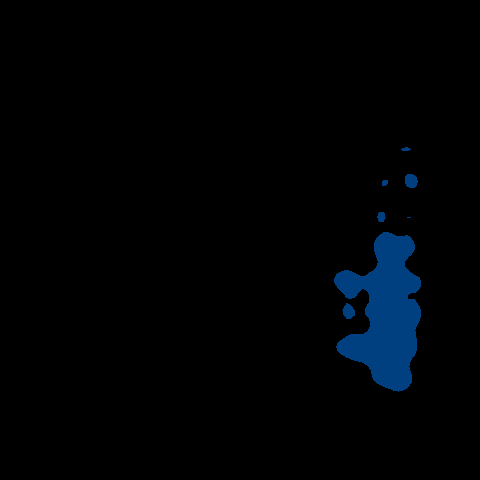}}\\

& \small{Clean} 
   & {\small{MI-FGSM~\cite{dong2018boosting}}}  &{\small{$\text{DI}^{2}\text{-FGSM}$~\cite{xie2019improving}}}  & {\small{ILA~\cite{huang2019enhancing}}}  & {\small{NAA~\cite{zhang2022improving}}} & {\small{RAP~\cite{qin2022boosting}}}  & {\small{Ours}}  \\
   \end{tabular} 
    \caption{Visual comparison of transfer-based black-box attacks on semantic segmentation task with DL3Mob as the surrogate model backbone. The first column represents the clean images and the corresponding predictions.} 
    \label{fig:transfer_comparison_mobilenet}
\end{figure*}

\begin{figure*}[tph]
   \centering
   \begin{tabular}{{c@{ } c@{ } c@{ } c@{ } c@{ }c@{ } c@{ } c@{ }c@{ } }}
 \rotatebox{90}{~~~ \small Image}& 
 {\includegraphics[width=0.13\linewidth,height=0.10\linewidth]{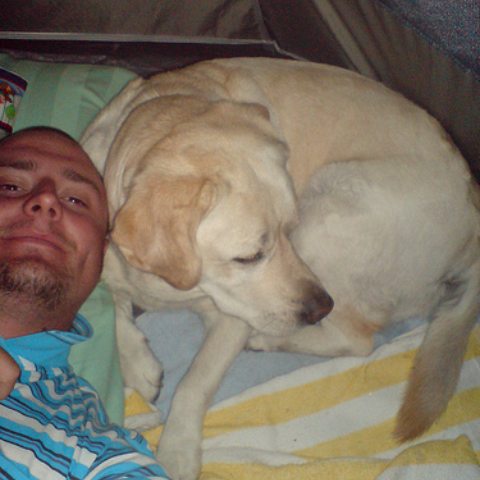}}&
 {\includegraphics[width=0.13\linewidth,height=0.10\linewidth]{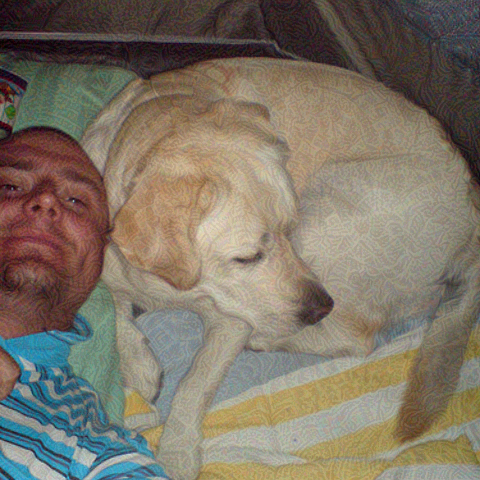}}&
 {\includegraphics[width=0.13\linewidth,height=0.10\linewidth]{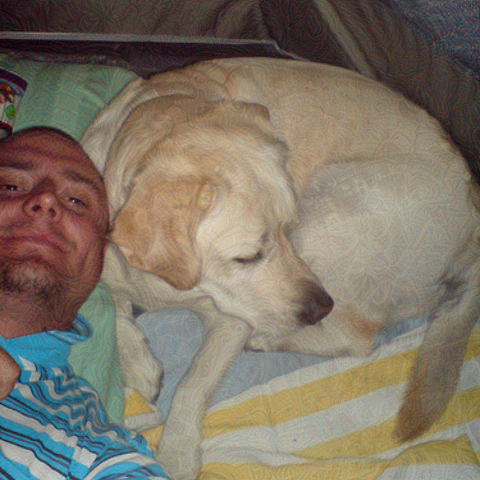}}&
 {\includegraphics[width=0.13\linewidth,height=0.10\linewidth]{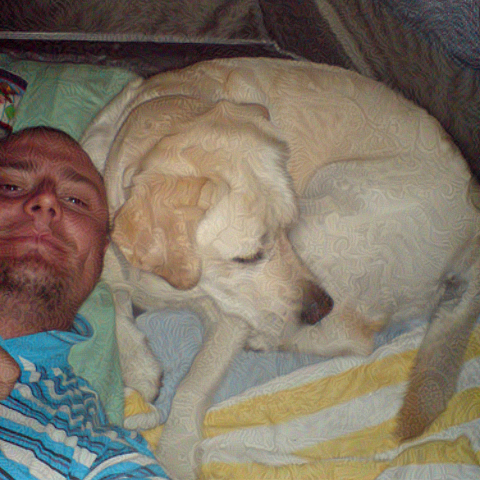}}&
 {\includegraphics[width=0.13\linewidth,height=0.10\linewidth]{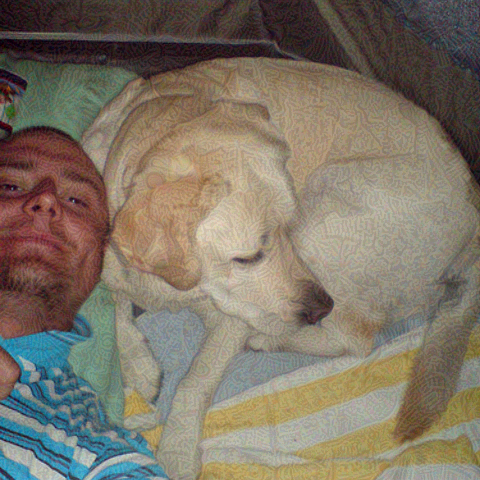}}&
 {\includegraphics[width=0.13\linewidth,height=0.10\linewidth]{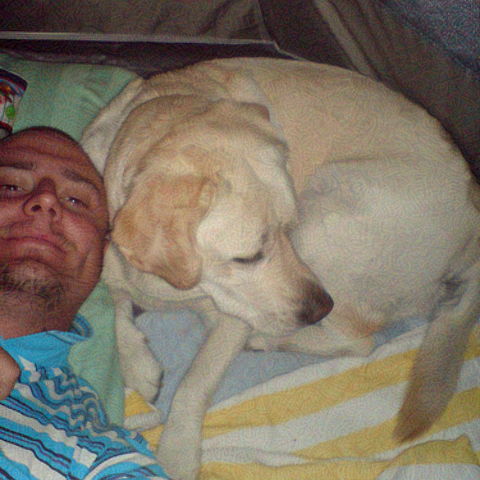}}&
 {\includegraphics[width=0.13\linewidth,height=0.10\linewidth]{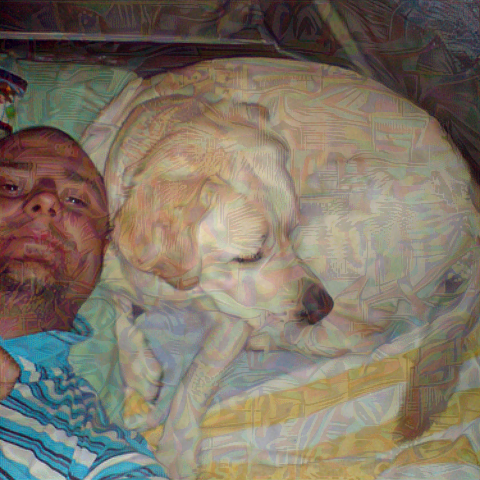}}
 \\
 \rotatebox{90}{~~ \small DL3Mob}& 
 {\includegraphics[width=0.13\linewidth,height=0.10\linewidth]{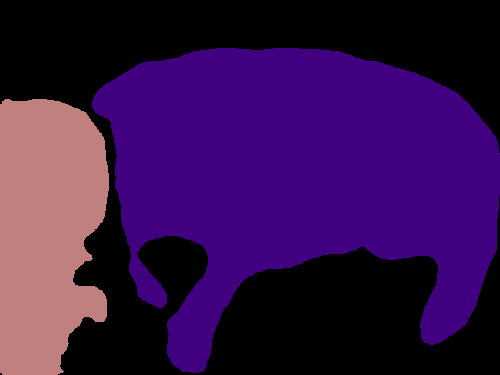}}&
 {\includegraphics[width=0.13\linewidth,height=0.10\linewidth]{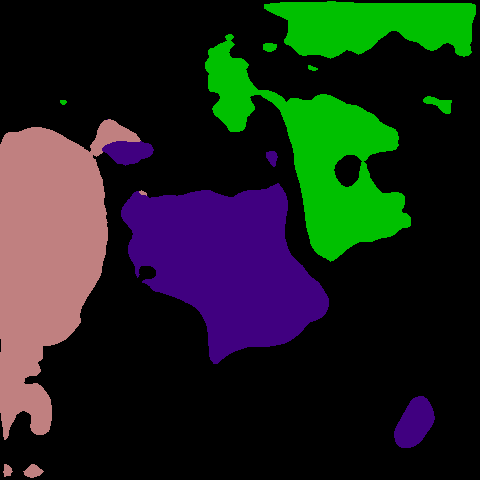}}&
 {\includegraphics[width=0.13\linewidth,height=0.10\linewidth]{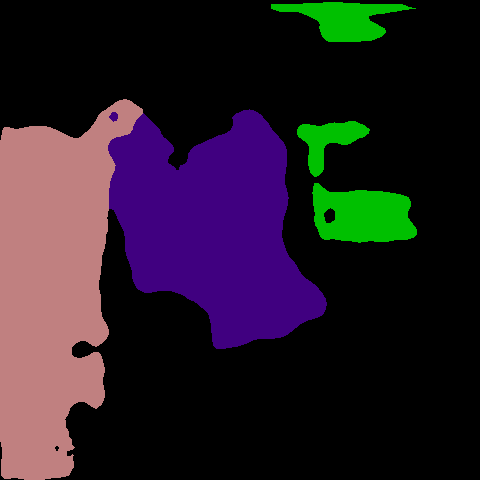}}&
 {\includegraphics[width=0.13\linewidth,height=0.10\linewidth]{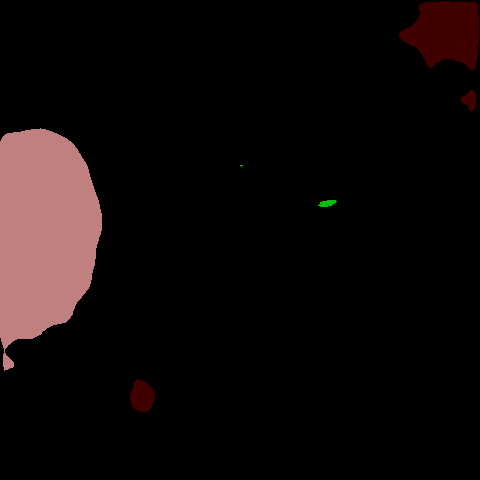}}&
 {\includegraphics[width=0.13\linewidth,height=0.10\linewidth]{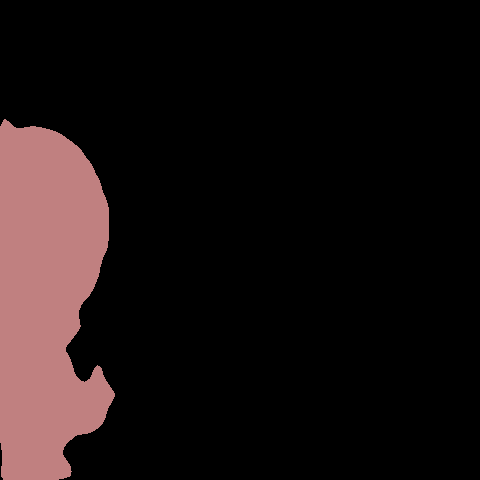}}&
 {\includegraphics[width=0.13\linewidth,height=0.10\linewidth]{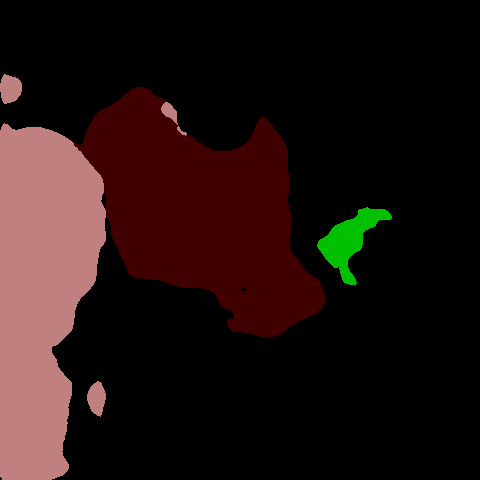}}&
 {\includegraphics[width=0.13\linewidth,height=0.10\linewidth]{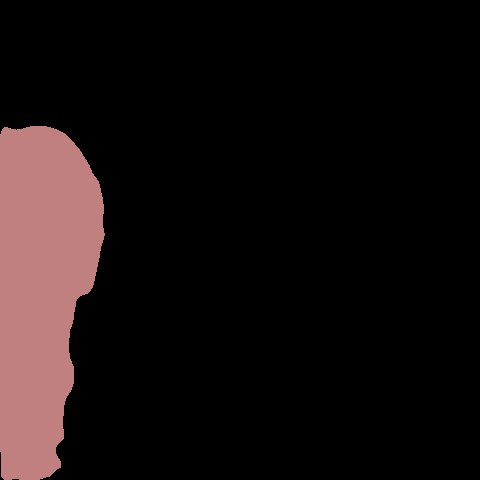}}
 \\
 \rotatebox{90}{~ \small DL3R101}& 
 {\includegraphics[width=0.13\linewidth,height=0.10\linewidth]{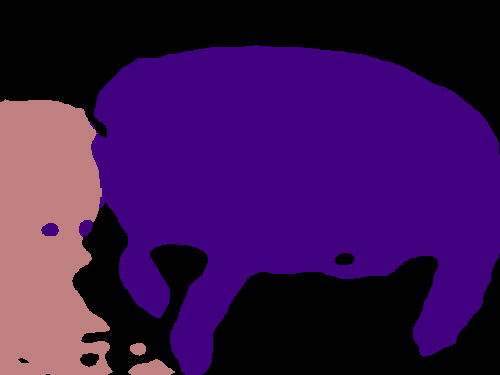}}&
 {\includegraphics[width=0.13\linewidth,height=0.10\linewidth]{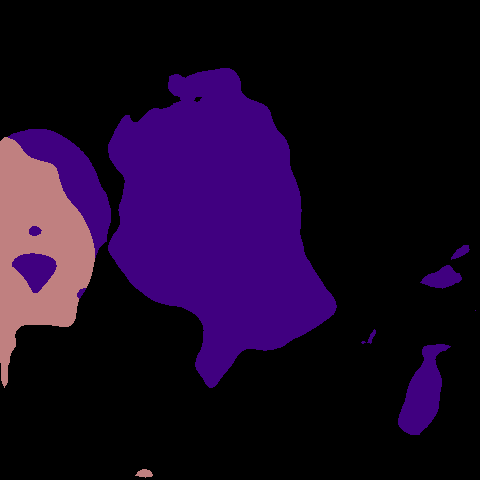}}&
 {\includegraphics[width=0.13\linewidth,height=0.10\linewidth]{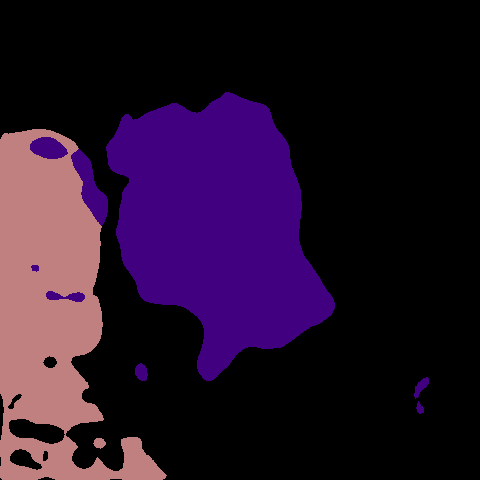}}&
 {\includegraphics[width=0.13\linewidth,height=0.10\linewidth]{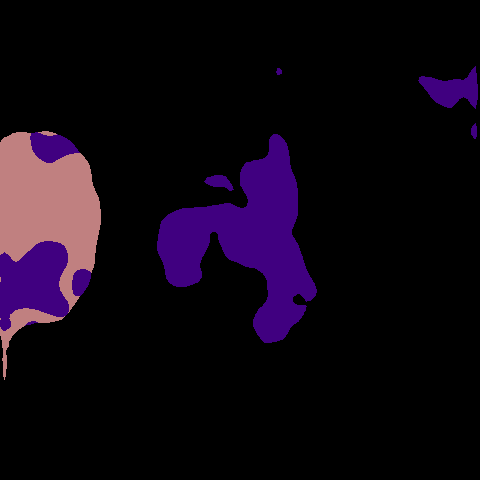}}&
 {\includegraphics[width=0.13\linewidth,height=0.10\linewidth]{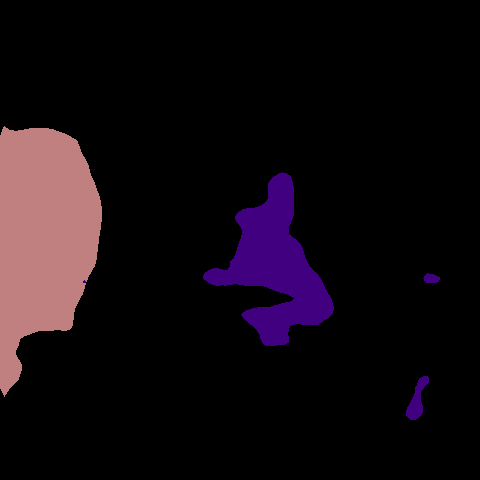}}&
 {\includegraphics[width=0.13\linewidth,height=0.10\linewidth]{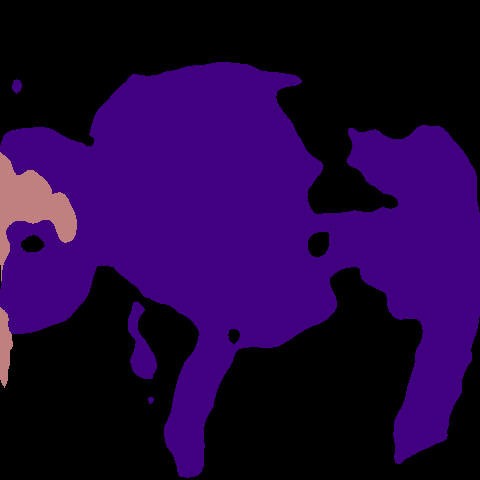}}&
 {\includegraphics[width=0.13\linewidth,height=0.10\linewidth]{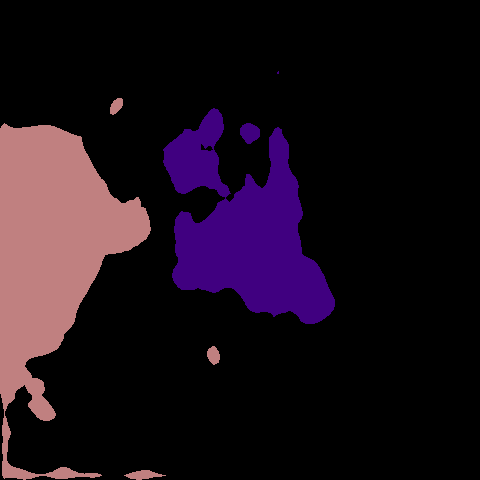}}\\

& \small{Clean} 
   & {\small{MI-FGSM~\cite{dong2018boosting}}}  &{\small{$\text{DI}^{2}\text{-FGSM}$~\cite{xie2019improving}}}  & {\small{ILA~\cite{huang2019enhancing}}}  & {\small{NAA~\cite{zhang2022improving}}} & {\small{RAP~\cite{qin2022boosting}}}  & {\small{Ours}}  \\
   \end{tabular} 
    \caption{Visual comparison of transfer-based black-box attacks on semantic segmentation task with PSPR50 as the surrogate model backbone. The first column represents the clean images and the corresponding predictions.} 
    \label{fig:transfer_comparison_psp}
\end{figure*}

\textit{For COD task,} the results for transfer-based black-box attack methods using ViT and ResNet50 as the surrogate model are shown in \cref{tab:transfer_compare_dpt} and \cref{tab:transfer_compare_R50}, respectively. 
It can be observed that our proposed data distribution-based attack algorithm successfully circumvents the restriction of confining the adversarial samples to the structure of the surrogate model, generating more transferable adversarial perturbations.

\textit{For the semantic segmentation task,} the results for transfer-based black-box attack methods using PSPR50 or DL3MOB as the surrogate model is shown in \cref{tab:transfer_compare_segment_psp}, where we simultaneously compared the transferability of adversarial samples generated by PGD~\cite{madry2017towards} and SegPGD~\cite{gu2022segpgd} on the PSPR50 model across various networks. We observed that, in semantic segmentation tasks, white-box attacks exhibit similar transferability to transfer-based black-box attacks, consistent with the observations in \cite{gu2021adversarial}.
Our method's advantage over others lies in its ability to generate adversarial samples based on data distribution without requiring a task-specific victim model. \cref{tab:transfer_compare_segment_psp} demonstrates that our method can produce transferable adversarial samples for semantic segmentation. However, one limitation of our method is its inability in searching for the most vulnerable misclassification errors as easily as gradient-based approaches relying on surrogate model loss in multi-class tasks~(multi-peak problem). The main reason is that perturbation from our method is not directly generated from misclassification errors, but derived from scores obtained from score based models. A deep correlation exploration between classification error and score will be investigated further.

\begin{figure*}[!hpt]
   \centering
 \adjustboxset{width=0.130,height=0.10,valign=c}
   \begin{tabular}{{c@{ } c@{ } c@{ } c@{ } c@{ }c@{ } c@{ } c@{ } }}
   \multirow{2}*[2mm]{\rotatebox[origin=c]{90}{\small Swin}}&
 {\includegraphics[width=0.130\linewidth,height=0.10\linewidth]{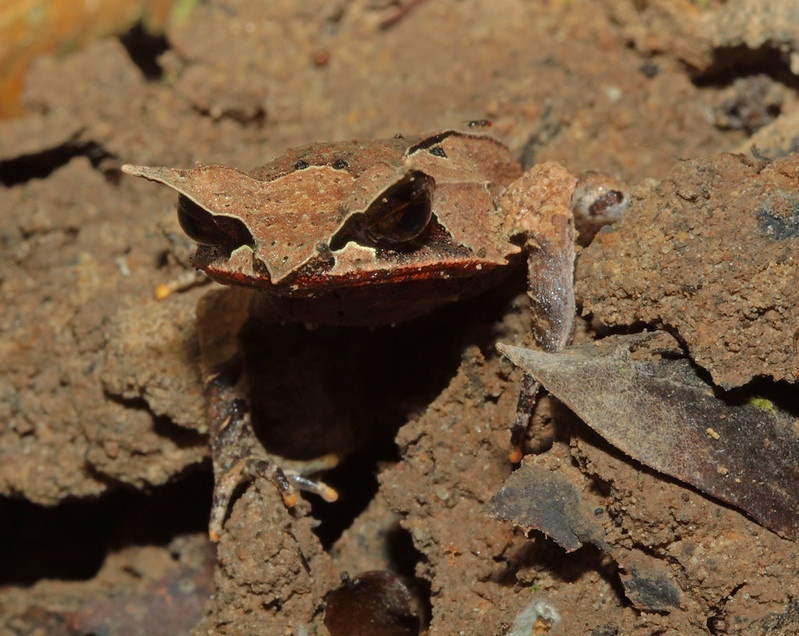}}&
 {\includegraphics[width=0.130\linewidth,height=0.10\linewidth]{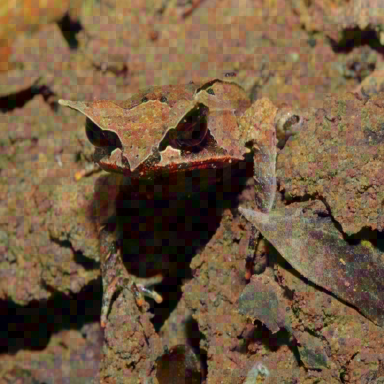}}&  
 {\includegraphics[width=0.130\linewidth,height=0.10\linewidth]{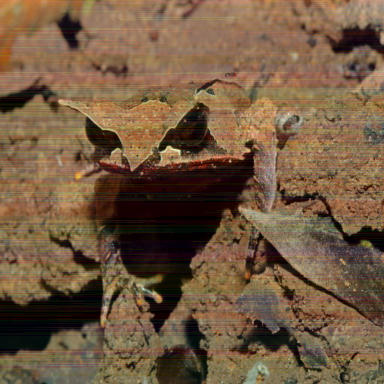}}&  
 {\includegraphics[width=0.130\linewidth,height=0.10\linewidth]{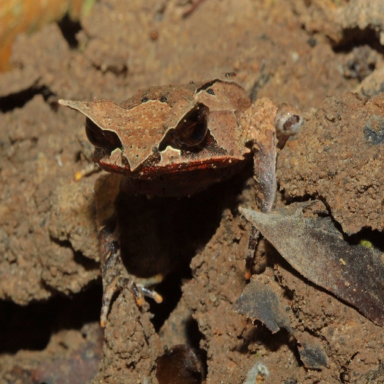}}&  
 {\includegraphics[width=0.130\linewidth,height=0.10\linewidth]{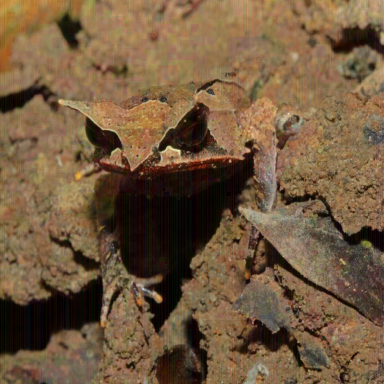}}&  
 {\includegraphics[width=0.130\linewidth,height=0.10\linewidth]{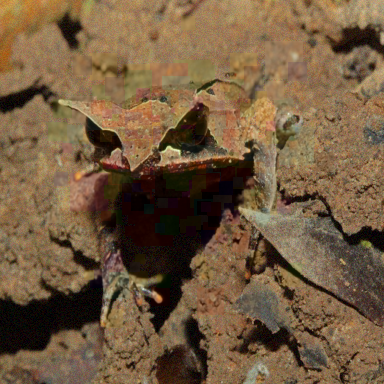}}&  
 {\includegraphics[width=0.130\linewidth,height=0.10\linewidth]{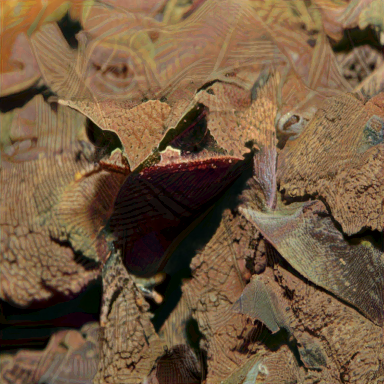}}
 \\
 & {\includegraphics[width=0.130\linewidth,height=0.10\linewidth]{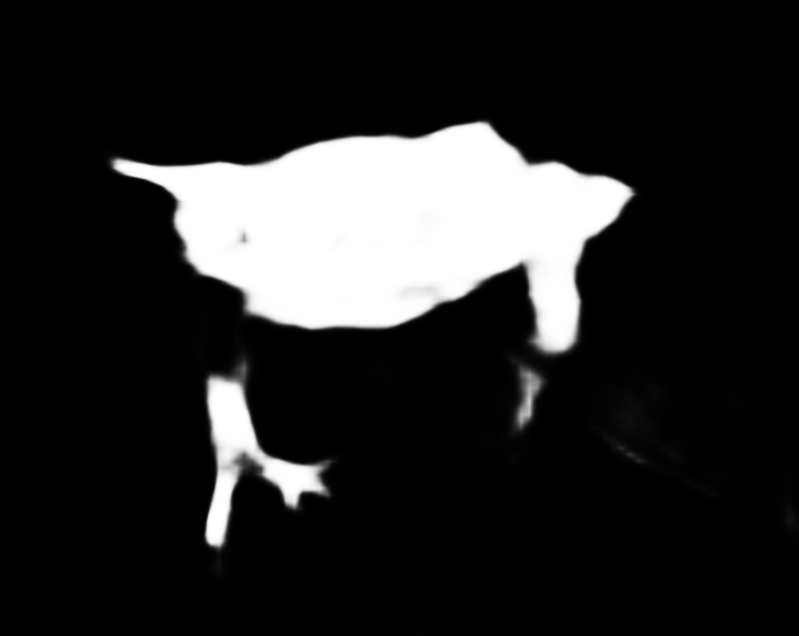}}&  
{\includegraphics[width=0.130\linewidth,height=0.10\linewidth]{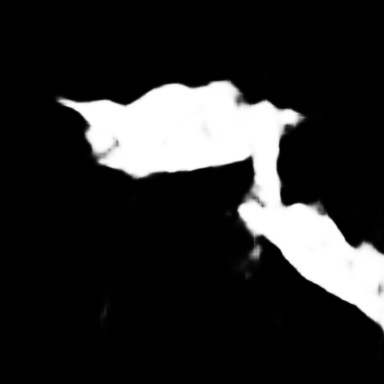}}&  
{\includegraphics[width=0.130\linewidth,height=0.10\linewidth]{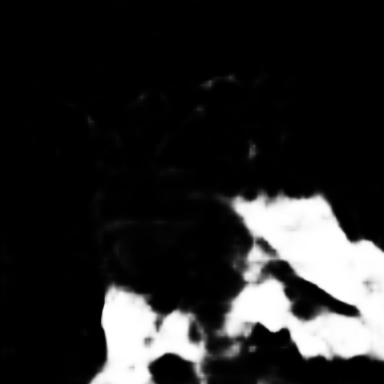}}&  
{\includegraphics[width=0.130\linewidth,height=0.10\linewidth]{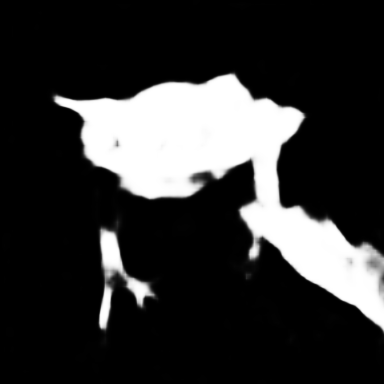}}& 
{\includegraphics[width=0.130\linewidth,height=0.10\linewidth]{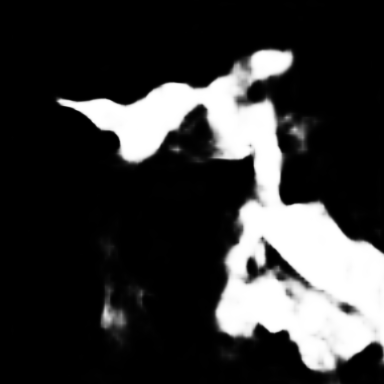}}& 
{\includegraphics[width=0.130\linewidth,height=0.10\linewidth]{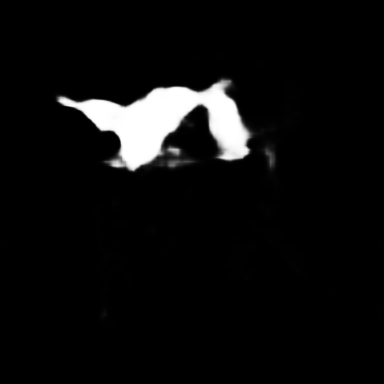}}&  
{\includegraphics[width=0.130\linewidth,height=0.10\linewidth]{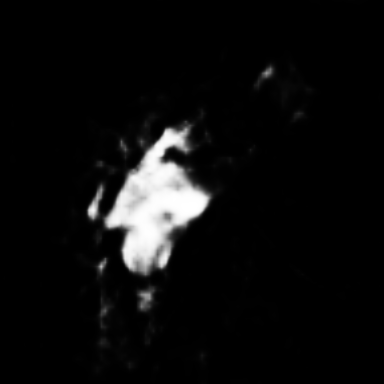}} \\
    \multirow{2}*[3mm]{\rotatebox[origin=c]{90}{\small PVTv2}}&
 {\includegraphics[width=0.130\linewidth,height=0.10\linewidth]{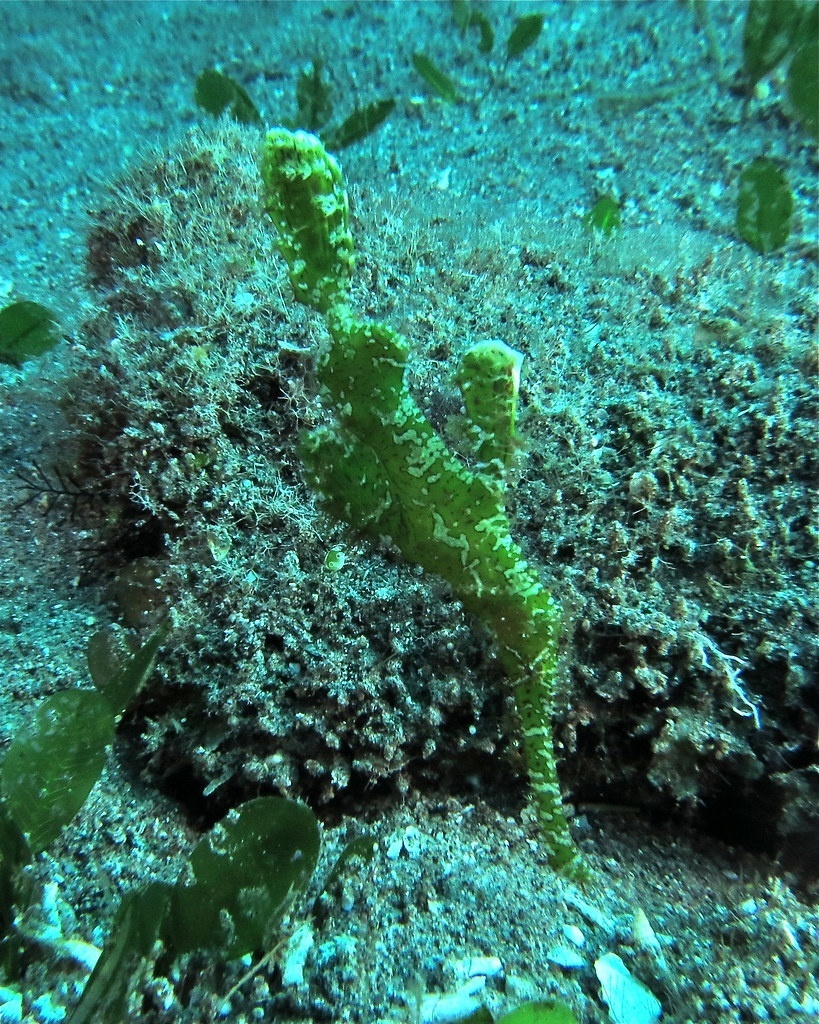}}&
 {\includegraphics[width=0.130\linewidth,height=0.10\linewidth]{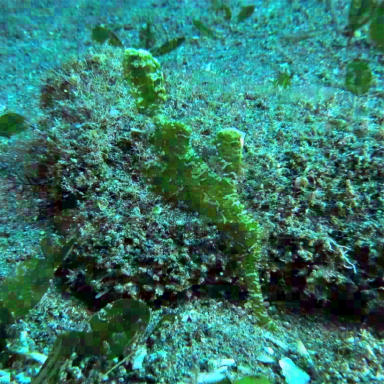}}&  
 {\includegraphics[width=0.130\linewidth,height=0.10\linewidth]{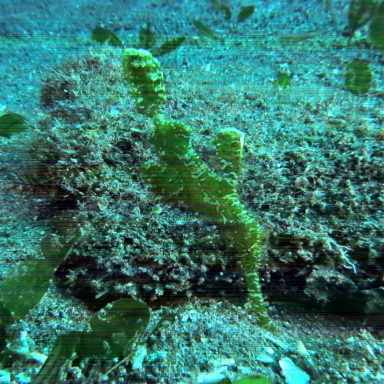}}&  
 {\includegraphics[width=0.130\linewidth,height=0.10\linewidth]{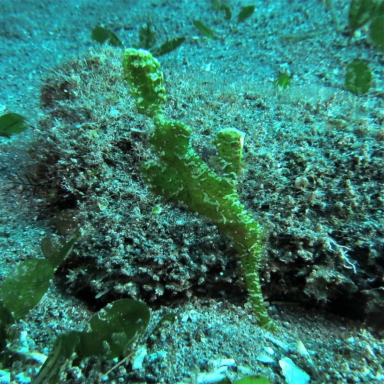}}&  
 {\includegraphics[width=0.130\linewidth,height=0.10\linewidth]{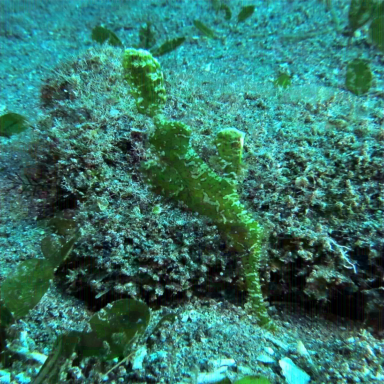}}&  
 {\includegraphics[width=0.130\linewidth,height=0.10\linewidth]{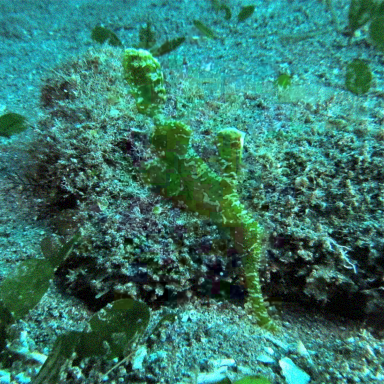}}&  
 {\includegraphics[width=0.130\linewidth,height=0.10\linewidth]{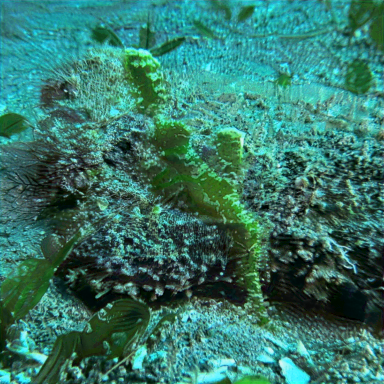}}
 \\
 & {\includegraphics[width=0.130\linewidth,height=0.10\linewidth]{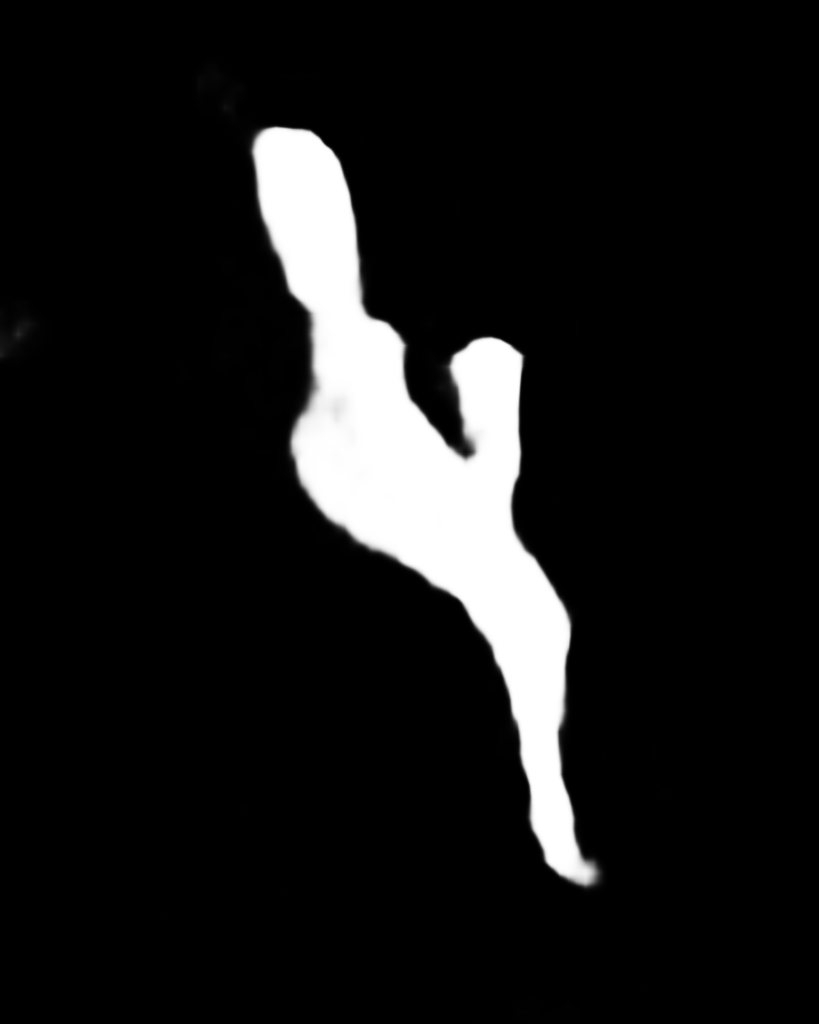}}&  
{\includegraphics[width=0.130\linewidth,height=0.10\linewidth]{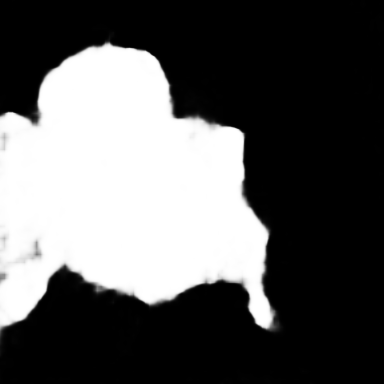}}&  
{\includegraphics[width=0.130\linewidth,height=0.10\linewidth]{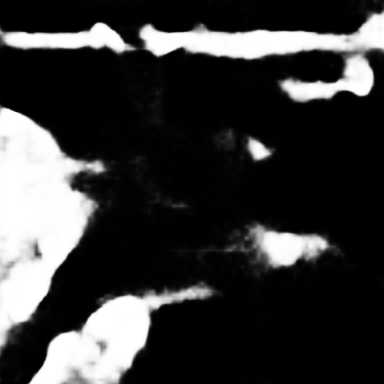}}&  
{\includegraphics[width=0.130\linewidth,height=0.10\linewidth]{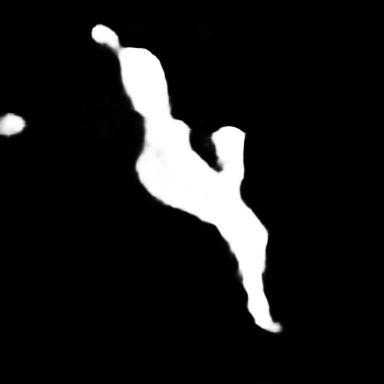}}& 
{\includegraphics[width=0.130\linewidth,height=0.10\linewidth]{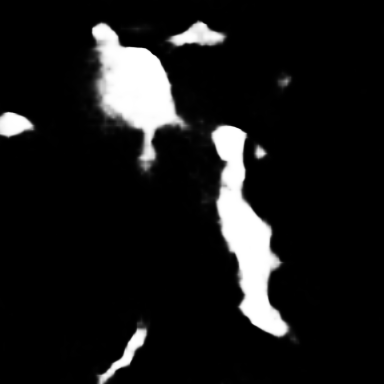}}& 
{\includegraphics[width=0.130\linewidth,height=0.10\linewidth]{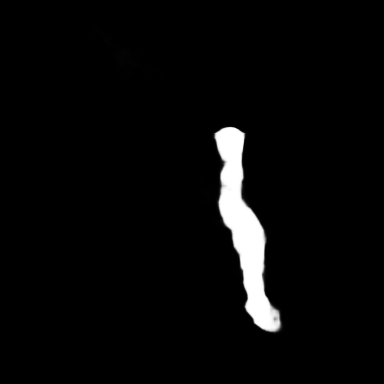}}&  
{\includegraphics[width=0.130\linewidth,height=0.10\linewidth]{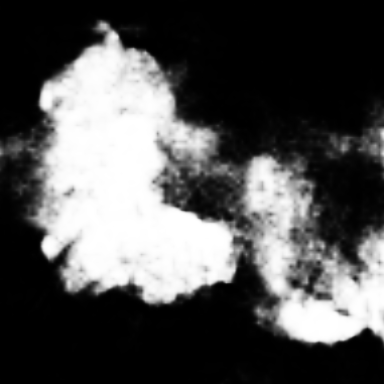}} \\
   & \small{Clean} 
   & {\small{Bandit~\cite{ilyas2018prior}}}  &{\small{{SignHunter~\cite{al2019sign}}}}  & {\small{SimBA~\cite{guo2019simple}}}  & {\small{Square~\cite{andriushchenko2020square}}} & {\small{IASA~\cite{li2022query}}} & {\small{Ours}}  \\
   \end{tabular} 
    \caption{Visual comparison with query-based black-box attacks on COD task, where each paired lines depict the adversarial samples (top) and the predictions (bottom) of victim models with Swin and PVTv2 backbones. }
    \label{fig:query_comparison}
\end{figure*}
\begin{table*}[!hpt]
  \centering
  \small
  \renewcommand{\arraystretch}{1.10}
  \renewcommand{\tabcolsep}{0.2mm}
  \caption{Performance comparison with query-based black-box attacks and a universal attack on COD task. The term \enquote{t} denotes the minutes required to generate an adversarial sample.  Specifically, SimBA~\cite{guo2019simple} performs 10k attack iterations with about 17k queries per sample.}
 \resizebox{\linewidth}{!}{ \begin{tabular}{l|c|ccccc|ccccc|ccccc|ccccc|ccccc} 
 \bottomrule[1pt]
Victim &  & \multicolumn{5}{c|}{ViT~\cite{ranftl2021vision}}& \multicolumn{5}{c|}{PVTv2~\cite{wang2022pvt}}& \multicolumn{5}{c|}{R50~\cite{he2016deep}}  &  \multicolumn{5}{c|}{Swin~\cite{liu2021swin}}  &  \multicolumn{5}{c}{Vgg~\cite{vgg_network}} \\   \hline
     & query & t &$\mathcal{M}\uparrow$ &$CC\downarrow$ & $S_{\alpha}\downarrow$&$E_{\xi}\downarrow$& 
     t&$\mathcal{M}\uparrow$ &$CC\downarrow$ & $S_{\alpha}\downarrow$&$E_{\xi}\downarrow$& t&$\mathcal{M}\uparrow$ &$CC\downarrow$ & $S_{\alpha}\downarrow$&$E_{\xi}\downarrow$& t&$\mathcal{M}\uparrow$ &$CC\downarrow$ & $S_{\alpha}\downarrow$&$E_{\xi}\downarrow$& t&$\mathcal{M}\uparrow$ &$CC\downarrow$ & $S_{\alpha}\downarrow$&$E_{\xi}\downarrow$\\
  \hline
Baseline  && \multicolumn{1}{c|}{-} & .026 & .781 & .852 & .924 & \multicolumn{1}{c|}{-} &.024 & .797 & .866 & .932 & \multicolumn{1}{c|}{-} &.041 & .681 & .792 & .864 & \multicolumn{1}{c|}{-} & .029 & .759 & .841 & .916 & \multicolumn{1}{c|}{-} &.048 & .644 & .764 & .836\\
Bandit~\cite{ilyas2018prior}  &10k &  \multicolumn{1}{c|}{10.7} &  .155 & .347 & .565 & .616 & \multicolumn{1}{c|}{26.0}& .150 & .388 & .590 & .629&  \multicolumn{1}{c|}{5.1} & .258 & .200 & .469 & .478 & \multicolumn{1}{c|}{11.1} & .135 & .364 & .582 & .639 &  \multicolumn{1}{c|}{7.2} & .350 & .120 & .379 & .404 \\
SignHunter~\cite{al2019sign}  &10k &\multicolumn{1}{c|}{22.5} & .304 & .190 & .423 & .464 &  \multicolumn{1}{c|}{26.2} & .265 & .228 & .460 & .495&  \multicolumn{1}{c|}{5.3}& .616 & .088 & .227 & .256 &  \multicolumn{1}{c|}{11.6}& .286 & .198 & .434 & .484 &  \multicolumn{1}{c|}{7.2}& .622 & .062 & .205 & .240 \\
SimBA~\cite{guo2019simple}  &17k & \multicolumn{1}{c|}{23.1}& .037 & .715 & .811 & .886& \multicolumn{1}{c|}{47.1} & .048 & .658 & .777 & .850 & \multicolumn{1}{c|}{16.6}& .072 & .541 & .706 & .777 & \multicolumn{1}{c|}{25.3} &.059 & .592 & .735 & .823 &\multicolumn{1}{c|}{14.3} & .087 & .469 & .661 & .732 \\  
Square~\cite{andriushchenko2020square}  &10k & \multicolumn{1}{c|}{5.1} &  .175 & .370 & .568 & .623 & \multicolumn{1}{c|}{13.7}& .176 & .386 & .579 & .620& \multicolumn{1}{c|}{2.7} & .443 & .162 & .352 & .382 &\multicolumn{1}{c|}{5.7} & .206 & .308 & .524 & .579 &\multicolumn{1}{c|}{3.8} & .435 & .128 & .334 & .368 \\
IASA~\cite{li2022query}  & 1k & \multicolumn{1}{c|}{0.5} &  .064 & .470 & .617 & .677&  \multicolumn{1}{c|}{1.2} &.057 & .530 & .652 & .726& \multicolumn{1}{c|}{0.3} & .089 & .227 & .512 & .551 & \multicolumn{1}{c|}{0.5}  &.068 & .406 & .592 & .655 & \multicolumn{1}{c|}{0.4}  &.098 & .105 & .467 & .469 \\
DUAP~\cite{zhang2021data} & - & \multicolumn{1}{c|}{-}  & .029 & .768 & .843 & .916 &  \multicolumn{1}{c|}{-}& .025 & .792 & .863 & .928 & \multicolumn{1}{c|}{-}& .049 & .633 & .758 & .827 &\multicolumn{1}{c|}{-} & .030 & .753 & .836 & .912 & \multicolumn{1}{c|}{-} &.072 & .503 & .673 & .742 \\ 
\hline
\textbf{Ours}\_query  &0.1k & \multicolumn{1}{c|}{0.3} &.085 & .513 & .673 & .759&  \multicolumn{1}{c|}{0.4} & .108 & .454 & .638 & .715 & \multicolumn{1}{c|}{0.3} &.145 & .323 & .567 & .633 & \multicolumn{1}{c|}{0.3} &.106 & .432 & .629 & .720 & \multicolumn{1}{c|}{0.3} & .186 & .245 & .505 & .582 \\ 
\toprule[1pt]
     \end{tabular}}
  \label{tab:query_compare}
\end{table*}
\subsubsection{Qualitative comparison:} 
\textit{For COD,}
we show predictions of various models for adversarial samples generated by the ViT backbone surrogate model using the transfer-based attack approach in \cref{fig:transfer_comparison}, showing that
the attacks generated by our method exhibit enhanced transferability across different models. 
\textit{For semantic segmentation,} we show the prediction results of various models for adversarial samples based on DL3Mob and  PSPR50 surrogate model using the transfer-based attack approach in \cref{fig:transfer_comparison_mobilenet} and \cref{fig:transfer_comparison_psp} respectively, further demonstrating superiority of our method.

\subsection{Ablation Study}
\subsubsection{Query Or Not}
Our method is capable of enhancing attack effectiveness through query-based techniques, as demonstrated in \cref{xgt_alg}.
We compare the performance of our proposed attack method on COD task with five query-based black-box attacks in \cref{tab:query_compare}. 
Bandit~\cite{ilyas2018prior} utilized a gradient prior to improve query efficiency. 
Simple Black-box Adversarial Attacks~(SignHunter)~\cite{al2019sign}  used binary sign flipping as an alternative to the gradient estimation process. 
SimBA~\cite{guo2019simple} enhanced searching efficiency by exploring perturbations in orthogonal spaces. 
Square Attack~(Square)~\cite{andriushchenko2020square} applied perturbations to squares at random locations. 
Improved Adaptive Square Attack~(IASA)~\cite{li2022query} learned the effect of square position on the attack based on \cite{andriushchenko2020square}.
We also compare to a data-free universal attack method, namely Data-free Universal Adversarial~(DUAP)~\cite{zhang2021data}, which attacks all segmentation models using a checkerboard-shaped perturbation. The number of queries is configured according to the original articles. We set $m_{\text{max}}=100$, and the query loss $L_{Q}$ in \cref{xgt_alg} is the binary cross-entropy loss.

The performance improvement from \cref{tab:transfer_compare_dpt} (\textbf{Ours}) to \cref{tab:query_compare} (\textbf{Ours}\_query) shows that our method provides the flexibility to choose whether or not to query the victim model and can significantly enhance attack effectiveness with a minimal number of query iterations. 
 Furthermore, it can even surpass the performance of certain query-based black-box attack algorithms. And the performance of our method without query is on par with the query-based black-box attack methods with 1000 queries, showing the potential of our new perspective. 
We showcase the predictions of the model with Swin and PVTv2 backbone for adversarial samples generated by query-based black-box attacks in \cref{tab:query_compare}. One observation is that our method is effective in generating adversarial samples in a small number of queries, causing different models to produce false predictions. 
Moreover, we find that SimBA~\cite{guo2019simple} has a poor attack performance, which we argue is caused by the fact that SimBA is attacking the camouflaged image in the frequency domain, however, both the camouflaged object and the background are at a lower frequency, so it is difficult for SimBA to locate the effective attack region of the camouflaged object.

\subsubsection{Selection of $\omega$} 
As in \cref{eq_conditional_seg_score_derivation_weighted}, $\omega$ is employed to mitigate the disparity between the actual score and the estimated score. 
Since the performance of query-free black-box attacks is related to the number of pre-defined attack steps, we compare the impact of different $\omega$ choices on query-based black-box attacks as shown in \cref{tab:omega_choice} with ResNet50 as the backbone for the victim model on COD task. \cref{tab:omega_choice} shows that our performance is relatively stable \wrt~the choice of $\omega$. We thus set $\omega=90$ to achieve a trade-off between efficiency and effectiveness.
\begin{table}[!h]
    \centering
        \caption{Performance \wrt~$\omega$, where $\textbf{B}$ denotes the baseline. }
\begin{tabular}{c|ccccccccc}
\bottomrule[0.75pt]
& \multicolumn{1}{c}{\textbf{B}}  & \multicolumn{1}{c}{\textbf{30}} & \multicolumn{1}{c}{\textbf{50}} & \multicolumn{1}{c}{\textbf{70}} & \multicolumn{1}{c}{\textbf{90}} & \multicolumn{1}{c}{\textbf{150}} & \multicolumn{1}{c}{\textbf{200}} \\ \hline
\textbf{$\mathcal{M}\uparrow$}  & .041& .138& .142& \textbf{.145}& \textbf{.145}& \textbf{.145}& .143\\
\textbf{$CC\downarrow$}         & .681&  .339& .329& .324& \textbf{.323}& \textbf{.323}& .324\\
\textbf{$S_{\alpha}\downarrow$} & .792&  .577& .569& \textbf{.567}&\textbf{.567}& \textbf{.567}& \textbf{.567}\\
\textbf{$E_{\xi}\downarrow$  }  & .864&  .645& .639& .634&\textbf{.633}& .635& .637\\ 
\toprule[0.75pt]
\end{tabular}
    \label{tab:omega_choice}
\end{table}

\subsubsection{Iteration of adversarial samples} 
In \cref{adv_1add_noise}, we emulate the noise addition process of the generative diffusion model by introducing adversarial perturbations into the sample $x^{\text{adv}}_{m}$.
The core idea is that the sample needs to simultaneously approximate the distribution during the diffusion training and meet the attack constraint,
$\ell_{\infty}<\delta_{d}$. We compare the attack results without adding perturbation noise on COD task in \cref{tab:abu_add_gaussian_or_not}, showing improved performance via
overlaying the estimated noise.
We attribute this improvement to the more accurate gradient of the data distribution after introducing noise similar to the diffusion model training phase.
We also compare the results obtained by applying 100 steps of random noise as a perturbation, as shown in \textbf{PG} in \cref{tab:abu_add_gaussian_or_not}, further explains the superiority of our solution in \cref{adv_1add_noise}.
\begin{table}[!h]
 \centering
        \caption{Verification of \cref{adv_1add_noise} with our query-based model.}
\begin{tabular}{l|cc|cccc} 
             \bottomrule[0.75pt]
             &noise & clip &$\mathcal{M}\uparrow$ &$CC\downarrow$ & $S_{\alpha}\downarrow$&$E_{\xi}\downarrow$ \\
              \hline
            \textbf{Baseline} & - & - & .041 & .681 & .792 & .864  \\ \hline
            \multirow{3}{*}{\textbf{Ours} }     & - & \checkmark & .093 & .450 & .651 & .729 \\
                                                 & \checkmark & - & .092 & .463 & .660 & .735 \\
                                                & \checkmark & \checkmark  &\textbf{.145} & \textbf{.323} & \textbf{.567} & \textbf{.633}\\ \hline
            \textbf{PG}                   & Gs & \checkmark  &  .046 & .653 & .773 & .846 \\
                 \toprule[0.75pt]
     \end{tabular}
 \label{tab:abu_add_gaussian_or_not}
\end{table}

\subsubsection{Selection of $m_\text{max}$}
Without querying, the number of attack steps $m_\text{max}$ is an important factor for our attack. We thus compare the results of different attack steps in \cref{tab:abu_selection_of_tmax} with ResNet50 as the backbone for the victim model on COD task. The experiment proves that the results of different attack steps have fluctuations but the attack works effectively in general.

\begin{table}[!hpt]
  \centering
  \caption{Performance \wrt  $m_\text{max}$ without querying.}
 \begin{tabular}{l|c|cccc} 
 \bottomrule[0.75pt]
 &$m_{\text{max}}$ &$\mathcal{M}\uparrow$ &$CC\downarrow$ & $S_{\alpha}\downarrow$&$E_{\xi}\downarrow$ \\
  \hline
\textbf{Baseline} & - & .041 & .681 & .792 & .864  \\   \hline
\multirow{5}{*}{\textbf{Ours} }      & 20 &  .117 & .377 & .604 & .684\\
                                     & 30 & \textbf{.123} & \textbf{.364} & \textbf{.596} & \textbf{.670} \\
                                     & 40 & \textbf{.123} & .369 & .597 & \textbf{.670} \\
                                     & 50 &  .120 & .378 & .603 & .675 \\
                                     & 60 & .115 & .388 & .610 & .683 \\
                                     & 70 & .110 & .401 & .619 & .694\\
     \toprule[0.75pt]
     \end{tabular}
  \label{tab:abu_selection_of_tmax}
\end{table}

\subsubsection{Attack On Robust Model}
Robust models refer to models trained to be relatively robust to input perturbation. Following conventional practice~\cite{tramer2017ensemble} with ensemble adversarial training to achieve a robust model, we assess the efficacy of our approach alongside transfer-based black-box attack methods in the context of the COD task. Initially, we establish a model with ResNet50 as the backbone, leveraging FGSM to generate adversarial examples on two distinct backbone models, namely Swin and VGG. Subsequently, we integrate these adversarial examples into the training regimen to fortify the model's robustness. To inspect the efficacy of transfer-based black-box attacks against robust networks, we employ the fortified ResNet50 model, evaluating attack performance by utilizing PVTv2 as a surrogate model for adversarial sample generation. Comparing existing methods with our approach under the robust ResNet50 model, as depicted in \cref{tab:robust_attack}, further underscores the superiority of our new black-box attack without a specific victim model. The results obtained from testing the robust model using the original images are labeled as Clean.

\begin{table}[!htp]
  \centering
  \caption{Performance of the adversarial attack method on the robust ResNet50 model for the COD task, employing PVTv2 as the surrogate model backbone. Clean represents the result of testing the robust model using the original image. }
 \begin{tabular}{l|cccc} 
 \bottomrule[1pt]
 &$\mathcal{M}\uparrow$ &$CC\downarrow$ & $S_{\alpha}\downarrow$&$E_{\xi}\downarrow$ \\
  \hline
Clean & .049 & .627 & .749 & .819  \\ 
MI-FGSM~\cite{dong2018boosting} &  .063 & .543 & .705 & .777 \\ 
$\text{DI}^{2}\text{-FGSM}$~\cite{xie2019improving} & .051 & .617 & .747 & .816  \\ 
ILA~\cite{huang2019enhancing} &  .057 & .578 & .720 & .789 \\ 
NAA~\cite{zhang2022improving} & .053 & .606 & .737 & .804  \\ 
RAP~\cite{qin2022boosting} & .056 & .574 & .716 & .789  \\ 
\textbf{Ours}&  \textbf{.074} & \textbf{.483} & \textbf{.665} & \textbf{.739} \\
     \toprule[1pt]
     \end{tabular}
      \label{tab:robust_attack}
   
\end{table}

\subsubsection{Attack On Salient Object Detection Task}
In order to further validate the applicability of the proposed adversarial attack algorithm, we conducted additional experiments on the task of salient object detection~(SOD) to assess the attack performance of our proposed algorithm.
\noindent\textbf{Dataset:} Black-box attacks in real-world scenarios only have access to the victim model predictions, with no knowledge of other information. To rigorously validate the effectiveness of the attack algorithm, we train the victim model and the local model on two different datasets.
We employ the DUTS training dataset~\cite{wang2017learning} for training victim models, comprising 10553 images.
Due to the unavailability of a suitable training set for the condition-based score estimation model to ensure data isolation between the victim model and the local model,  we aggregated the test sets SOD~\cite{sod_dataset}, DUT~\cite{Manifold-Ranking:CVPR-2013}, ECSSD~\cite{yan2013hierarchical}, PASCAL-S~\cite{pascal_s_dataset}, SOC~\cite{fan2018SOC} comprising 8128 images in total, to form the training set for score estimation model.
We then evaluate the attack performance using the DUTS testing dataset, providing a more robust validation of the algorithms' attack effectiveness in the presence of data isolation. The model and attack settings are consistent with the COD task. 
The results of our proposed attack method are shown in \cref{tab:sod_attack}, where \textbf{Ours}\_query denotes the result with 100 queries. \textbf{Ours} denotes the result without query and the attack steps $m_{\text{max}}=30$.  The model trained with clean samples is denoted as the Baseline. \cref{tab:sod_attack} illustrates that our proposed method also exhibits adversarial capabilities in the SOD task.
\begin{table}[!h]
  \centering
   \renewcommand{\arraystretch}{1.0}
  \renewcommand{\tabcolsep}{0.2mm}
  \caption{Performance of our proposed adversarial attack method on the SOD task.}
 \resizebox{\linewidth}{!}{
 \begin{tabular}{l|cc|cc|cc|cc|cc} 
 \bottomrule[1pt]
 Victim &  \multicolumn{2}{c|}{Vit\cite{ranftl2021vision}} & \multicolumn{2}{c|}{Pvtv2\cite{wang2022pvt}}& \multicolumn{2}{c|}{R50\cite{he2016deep}}  &  \multicolumn{2}{c|}{Swin\cite{liu2021swin}}  &  \multicolumn{2}{c}{Vgg\cite{vgg_network}} \\ \hline
     &$\mathcal{M}\uparrow$ &$Corr\downarrow$ &$\mathcal{M}\uparrow$ &$Corr\downarrow$&$\mathcal{M}\uparrow$ &$Corr\downarrow$&$\mathcal{M}\uparrow$ &$Corr\downarrow$&$\mathcal{M}\uparrow$ &$Corr\downarrow$\\
  \hline
Baseline  & .024 & .896 &  .024 & .895 &  .037 & .842  & .029 & .875 &   .045 & .811 \\
\textbf{Ours}  &.041 & .839 & .042 & .833  & .074 & .716  & .047 & .812 & .095 & .636 \\
\textbf{Ours}\_query  &.064 & .763  &.056 & .787  &.117 & .602 & .066 & .752 &.132 & .520 \\
\toprule[1pt]
      \end{tabular}
      }
  \label{tab:sod_attack}
\end{table}

\section{Conclusion}
We study adversarial attack generation from an image generation perspective. Our derivation of using a weighted linear combination of conditional and unconditional scores as a replacement for conditional segmentation score is significantly different from existing solutions. Additionally, our method offers the flexibility to incorporate queries to enhance attack performance. 
The experimental results demonstrate that our method can generate generalized adversarial samples without requiring task-specific victim models for both binary segmentation and multi-class segmentation tasks. Our approach achieved optimal performance in the camouflaged object detection task and demonstrated effectiveness in multi-class tasks. We notice one main limitation of our method is the implicit correlation between classification error and score estimation, leading to less effective performance for some scenarios.
Further study to correlate these to terms will be conducted to further explore the potential of our solution in multi-category settings.

\bibliographystyle{IEEEtran}
\bibliography{aaai25}

\ifCLASSOPTIONcaptionsoff
  \newpage
\fi

\end{document}